\documentclass{article} % For LaTeX2e
\usepackage{iclr2027_conference,times}

\usepackage{amsmath,amsfonts,amssymb,bm}

\def\eqref#1{equation~\ref{#1}}
\def\ceil#1{\lceil #1 \rceil}
\def\floor#1{\lfloor #1 \rfloor}
\def\1{\bm{1}}

\DeclareMathAlphabet{\mathsfit}{\encodingdefault}{\sfdefault}{m}{sl}
\SetMathAlphabet{\mathsfit}{bold}{\encodingdefault}{\sfdefault}{bx}{n}

\newcommand{\E}{\mathbb{E}}

\newcommand{\R}{\mathbb{R}}

\newcommand{\Var}{\mathrm{Var}}

\DeclareMathOperator{\Tr}{Tr}

\usepackage{graphicx}
\usepackage{subcaption}
\usepackage[colorlinks=true, allcolors=blue]{hyperref}
\usepackage[capitalize,noabbrev]{cleveref}
\usepackage{url}
\usepackage{titletoc}
\usepackage{placeins}
\usepackage{amsthm}
\usepackage{thmtools,thm-restate}
\newtheorem{theorem}{Theorem}
\newtheorem{lemma}{Lemma}

\title{Renormalized Entropy from Symmetries:\\ Simplicity Bias and Neural Heat Death}

\author{Vasco Portilheiro\thanks{Corresponding author. Email: \texttt{vportilheiro@gatsby.ucl.ac.uk}.}, Samuel Liebana, Andrew Saxe, Peter Orbanz \\
Gatsby Computational Neuroscience Unit\\
University College London
}

\iclrfinalcopy % Uncomment for camera-ready version, but NOT for submission.
\begin{document}

\maketitle

\begin{abstract}
Neural networks have parameter symmetries: parameter transformations that do not affect the represented function. How do these symmetries affect the functions that neural networks ultimately learn? We study the implicit bias arising from these symmetries by calculating the equilibrium distribution over represented functions under Langevin dynamics with Gaussian regularization. To isolate the effect of symmetries on this distribution, we study the limit in which the strength of the regularization is taken to zero. In certain cases, we show that the zero-regularization and infinite-time limits commute. In the low temperature limit, we recover known effective regularizations. At positive temperature, a renormalization procedure allows us to derive a novel entropy which predicts symmetry-induced preference for certain functions. For linear networks, we obtain explicit entropy terms that favor singular and low-rank functions. For networks with homogeneous activations, we find entropy terms that favor inactive neurons, which can lead to neural "heat death" in which equilibrium mass concentrates on networks with vanishing output. We interpret these results through the neural tangent kernel, as a simplicity bias selecting for functions that vary slowly under variations in the parameters.
\end{abstract}

\section{Introduction}
Neural networks generally exhibit \emph{parameter symmetry}, meaning there are directions in parameter space which do not affect the network's input-output map. Examples include rescaling in homogeneous neural networks, and changes of bases for matrix products in linear networks and attention, among others \citep{saxe2014exactsolutionsnonlineardynamics, kunin2021neuralmechanicssymmetrybroken, zhao2025symmetryneuralnetworkparameter}. How does training (which occurs in parameter space) affect what functions are actually learned? 

In general, we can describe the situation as follows. Let $\theta \in \Theta \subseteq \R^N$ be the neural network parameters and $f = \pi(\theta)$ be the represented function. Assume $\theta$ is optimized to minimize the loss $L(\theta)$. Since the loss should only depend on $f$, we write $L(\theta) = \ell(f)$. Optimization in parameter space induces non-trivial function space dynamics, for example for gradient flow (with rate $\eta>0$)
\begin{align*}
\dot\theta = -\eta \nabla L(\theta) \qquad \Longrightarrow \qquad \dot f = -\eta D\pi_\theta D\pi_\theta^\top \nabla\ell(f).
\end{align*}
Here, $D\pi_\theta$ denotes the Jacobian of $\pi$ at $\theta$, and $D\pi_\theta D\pi_\theta^\top$ is the neural tangent kernel \citep{jacot2018ntk}. The difficulty in analyzing implicit bias in $f$ dynamically is that the NTK evolves over time, and in general depends on $\theta$ in complicated ways. In certain cases this dependence can be simplified, as characterized by \citet{marcotte2025intrinsic}, or in various limits using sophisticated techniques to find order parameters \citep{bordelon2022selfconsistent}.

We consider a different approach, looking at the steady-state distribution $p(\theta)$ of parameters subjected to noisy training dynamics. This allows us to abstract away the time course of the dynamics and study the steady-state distribution they induce over functions,
\begin{align*}
\label{eq:cartoon-equilibrium}
\mathrm{steady\ state} = p(\theta) \qquad \Longrightarrow \qquad \Pr(f \in B) = \int_{\pi^{-1}(B)} p(\theta)\,d\theta.
\end{align*}
Neural network parameter steady state distributions exist in practice, though for SGD they are generally non-equilibrium (having probability currents) \citep{mandt2017sgd,chaudhari2018variational, yaida2019fluctuation,zhang2022invariant}, and may be complicated to write down analytically. A principled but tractable alternative is provided by Langevin dynamics, which can be thought of as gradient flow perturbed by white noise. While Langevin dynamics does not capture the anisotropic noise known to be important to the implicit bias of SGD \citep{zhu2019anisotropic,haochen2021shape}, it is nonetheless a classical model for studying Bayesian inference in neural networks \citep{welling2011sgld,mandt2017sgd} as well as the behavior of biological neural systems \citep{kappel2015plasticity,aitchison2021plasticity}. Crucially, Langevin dynamics' tractable Gibbs equilibrium allows us to make concrete predictions about the long-time symmetry-induced bias of otherwise complicated dynamics.

\begin{figure}[t]
    \centering
    \begin{subfigure}[t]{0.55\textwidth}
        \centering
        \includegraphics[width=\textwidth]{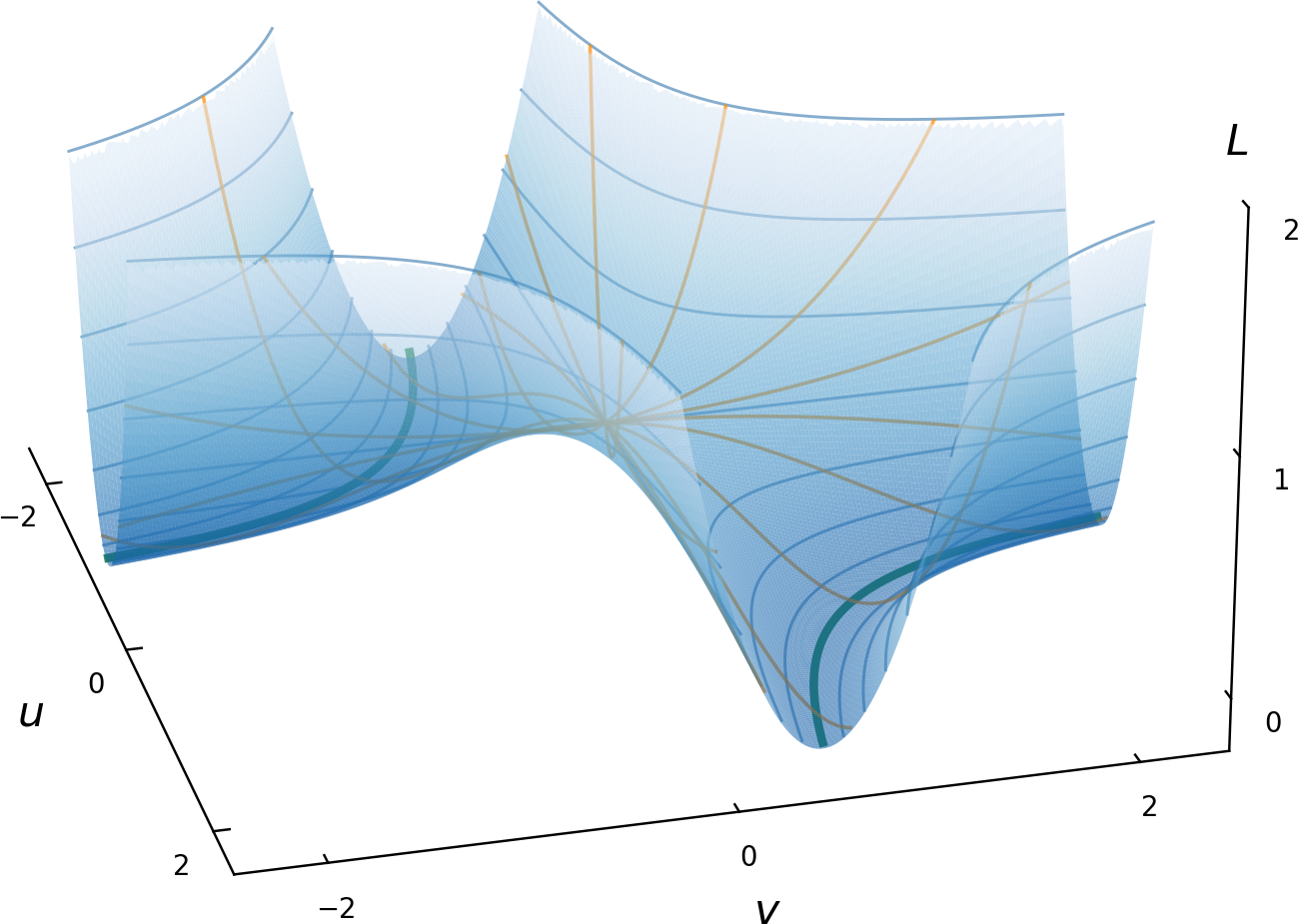}
        \caption{Surface view}
        \label{fig:diag_loss_surface}
    \end{subfigure}
    \hfill
    \begin{subfigure}[t]{0.42\textwidth}
        \centering
        \includegraphics[width=\textwidth]{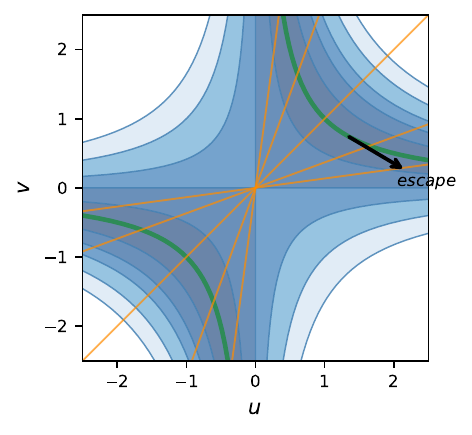}
        \caption{Contour view}
        \label{fig:diag_loss_contour}
    \end{subfigure}
    \caption{The loss landscape of $f(x)=uvx$ with $L(u,v)=(uv-1)^2/2$, plotted with the coordinates $w=uv$ (blue) and $a$ (orange, defined in \cref{sec:diag}). The loss and $w$ have scaling symmetry $u\mapsto cu$ and $v\mapsto v/c$. Under Langevin dynamics, $\theta$ escapes to infinity along $uv=1$ (green).}
    \label{fig:diag_loss}
\end{figure}

A central obstacle in our proposal is that the equilibrium distribution only exists when the loss landscape $L(\theta)$ prevents $\theta$ from diffusing ``to infinity.'' Common neural network symmetries include scaling, which is ``infinite'' (technically, \emph{non-compact}): you can always scale a number to make it bigger. Since scaling is a symmetry of the minimum-loss set, $\theta$ can escape along this set, meaning no stationary distribution exists (\cref{fig:diag_loss}). To circumvent this we consider squared-norm regularization (a choice we argue is well suited to Langevin, see \cref{sec:diag}) with strength $\lambda>0$,
\begin{align*}
L(\theta) + \lambda R(\theta), \qquad R(\theta)=\frac{1}{2}\|\theta\|^2.
\end{align*}
The regularization ``pulls up'' the loss landscape further from the origin, eliminating the possibility of escape and ensuring a stationary distribution exists. Such distributions are studied as Bayesian priors over functions \citep{li2021renormalization,izmailov2021posteriors,zavatoneveth2021exact,noci2021prior,kobialka2026priors}. Regularization, however, necessarily breaks the non-compact symmetries whose effect we wish to study. We therefore study the limit in which $\lambda$ is taken to zero. If the temperature in Langevin dynamics (defined below) is taken to zero first, we recover predictions from the representation cost literature \citep{srebro2004maxmargin,gunasekar2017matrix,zhen2021representationcost}, corresponding to selecting $f$ which minimizes $R$ subject to minimizing $\ell$. If temperature is instead fixed to be positive, using a renormalization procedure, we obtain a novel entropy-driven selection principle.

Our contributions in this paper are:
\begin{itemize}
    \item a conceptually simple but powerful framework for analyzing parameter symmetry, recovering known implicit biases in one limit, and a novel renormalized entropy in another;
    \item theoretical derivations of the renormalized entropy for diagonal, linear, and homogeneous networks, including the dominance of the zero-function for ReLU networks trained on radially coercive losses, which we call \emph{heat death};
    \item a proof that the $\lambda\to0$ and the long-time limits of unregularized Langevin dynamics correspond in diagonal networks, and simulations investigating whether these results generalize to other models;
    \item an interpretation of the entropy as selecting simple functions, with low NTK volume.
\end{itemize}

To the best of our knowledge, our work contains the first derivation of entropic implicit bias from the renormalized volume growth of non-compact symmetry fibers.

\section{Theoretical framework}
\label{sec:theory}

\paragraph{Langevin dynamics} Following the introduction, let $\theta \in \Theta \subseteq \R^N$ be neural network parameters, $f= \pi(\theta)$ and $L(\theta)=\ell(f)$. For a potential $U(\theta)$ and temperature $T>0$, Langevin dynamics is
\begin{align*}
d\theta_t = -\nabla U(\theta_t)\,dt + \sqrt{2T} dB_t
\end{align*}
where $B_t$ is $N$-dimensional Brownian motion. 
Under confinement and normalizability conditions such as, (1) $\langle \theta,\nabla U(\theta)\rangle \to \infty$ as $\|\theta\| \to \infty$, and (2) $\int \exp(-U(\theta)/T)\,d\theta$ is finite, there exists the Gibbs stationary distribution $p(\theta)\,d\theta \propto \exp(-U(\theta)/T)\,d\theta$.
This is the limiting distribution of $\theta_t$; that is if $p_t(\theta)$ is the probability density of $\theta_t$, then $p(\theta) = \lim_{t\to\infty}p_t(\theta)$. For losses $L(\theta)$ such as those shown in \cref{fig:diag_loss}, the normalization integral is not finite, so we let $U(\theta) = L(\theta) + \lambda R(\theta)$ and write the stationary distribution
\begin{align*}
p_{\lambda,T}(\theta)\,d\theta = \frac{e^{-\frac{1}{T}(L(\theta) + \lambda R(\theta))}}{\int e^{-\frac{1}{T}(L(\tilde\theta) + \lambda R(\tilde\theta))}\,d\tilde\theta}\,d\theta.
\end{align*}
The set $\pi^{-1}(f)=\{\theta\in\Theta:f=\pi(\theta)\}$ of parameters implementing the same function is called the \emph{fiber over $f$}. Since $L(\theta)=\ell(f)$, integrating over the fiber gives the induced distribution $f$
\begin{align*}
\Pr\nolimits_{\lambda,T}(df) \propto e^{-\ell(f)/T} \nu_{\lambda/T}(df),
\qquad
\nu_\alpha(B)=\int_{\pi^{-1}(B)}e^{-\alpha R(\theta)}\,d\theta.
\end{align*}
The probability distribution over $f$ is the Gibbs distribution for loss $\ell$, tilted by a factor $\nu_{\lambda/T}$ depending on the parameter symmetries encoded in $\pi$ --- counting representations of $f$ exponentially down-weighted by how much they cost in regularization.\footnote{
Having $R(\theta)=\|\theta\|^2/2$ as the regularization, $\exp(-\alpha R(\theta))$ is proportional to a Gaussian prior of precision (inverse variance) $\alpha$. Thus $\nu_\alpha(df)$ is proportional to the distribution of $\pi(\alpha^{-1/2} \xi)$ for $\xi\sim\mathcal{N}(0,I_N)$.
}
Studying this factor $\nu_{\lambda/T}$ for $\lambda\approx0$ reveals the implicit bias of the parameterization. However, for parameterizations with fibers extending to infinity (such as in \cref{fig:diag_loss}) $\nu_{0/T}$ may be infinite, meaning we must perform a limiting analysis.

\paragraph{Symmetry coordinates}{On a \emph{regular region} (or \emph{stratum}) of parameter space, where the Jacobian $D\pi$ has constant rank, we can locally define \emph{symmetry coordinates} $(f,a)$ (formally, a \emph{chart}), where $a$ parametrizes (a local part of) the fiber $\pi^{-1}(f)$ (see \cref{fig:diag_chart}). Writing the Jacobian of the change of variables as $d\theta = |J(f,a)|\,df\,da$, on such a region, we define the \emph{fiber partition function}}
\begin{align*}
Z_\alpha(f) = \int e^{-\alpha R(f,a)}|J(f,a)|\,da,
\end{align*}
where with an abuse of notation $R(f,a) = R(\theta)$. Then $\nu_\alpha(df)=Z_\alpha(f)\,df$, and we can write a density over functions,
\begin{align*}
p_{\lambda,T}(f) \propto e^{-\ell(f)/T} Z_{\lambda/T}(f).
\end{align*}

\paragraph{Zero-regularization limit}{We consider two regimes in which to take $\lambda\to0$. The simpler case is to take $T\to0$ first. For the fiber free energy $F_\alpha(f) = -\alpha^{-1}\log Z_\alpha(f)$, Laplace's principle gives}
\begin{align*}
\lim_{\alpha\to \infty} F_{\alpha}(f) = \inf_a R(f,a) =: R_\text{eff}(f),
\qquad
p_{\lambda,T}(f) \propto e^{-\frac{1}{T}(\ell(f) + \lambda F_{\lambda/T}(f))}.
\end{align*}
We see that taking $T\to0$ concentrates the distribution on the minimizers of $\ell(f)+\lambda R_\text{eff}(f)$. Taking $\lambda\to0$ next selects the minimizers of $\ell(f)$ of smallest effective regularization $R_\text{eff}(f)$.

We also consider fixing $T>0$ and taking $\lambda\to0$ directly. If a fiber $\pi^{-1}(f)$ had finite volume $\Omega(f)=\int |J(f,a)|\,da$, then $Z_{\alpha}(f)$ would converge to said volume, and we would have an entropic correction, $p_{0,T}(f)\propto \exp(-\ell(f)/T+\log \Omega(f))$. In general, however, $Z_\alpha(f)$ diverges as $\alpha\to0$. But suppose that on each regular region this divergence occurs at a well-defined rate, meaning
\begin{align*}
    \Omega(f) = \lim_{\alpha\to0} \frac{Z_\alpha(f)}{C_\alpha}
    \quad
    \text{exists for some }C_\alpha.
\end{align*}
Then assuming the limit can be passed through the normalization integral over $f$, and no other stratum contributes, we get
\begin{align*}
\lim_{\lambda\to0}p_{\lambda,T}(f) \propto e^{-\ell(f)/T + \log\Omega(f)}.
\end{align*}
In this case, we call $\log \Omega(f)$ the \emph{renormalized entropy} on the chosen stratum: $\Omega(f)$ is the relative volume of $\pi^{-1}(f)$ after removing a common divergence. Different strata may have different divergent $C_\alpha$, and the one with fastest growth as $\alpha\to0$ dominates the limiting probability, as we demonstrate below for ReLU networks. Note that $\Omega(f)$ can depend on the choice of regularization. We argue below that squared-norm is natural here because it matches the Gaussian diffusion. 

\section{Diagonal networks}
\label{sec:diag}
\begin{figure}[t]
    \centering
    \begin{subfigure}[t]{0.396\textwidth}
        \centering
        \includegraphics[width=\textwidth]{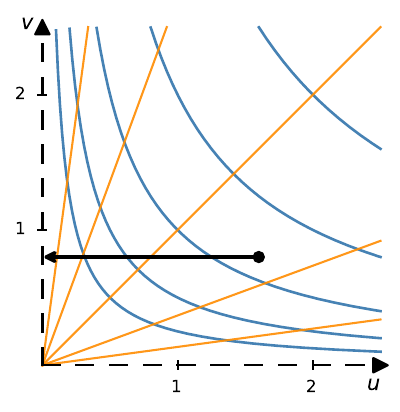}
        \caption{$u>0, v>0$ quadrant}
        \label{fig:uv_quadrant}
    \end{subfigure}
    \hfill
    \begin{subfigure}[t]{0.45\textwidth}
        \centering
        \includegraphics[width=\textwidth]{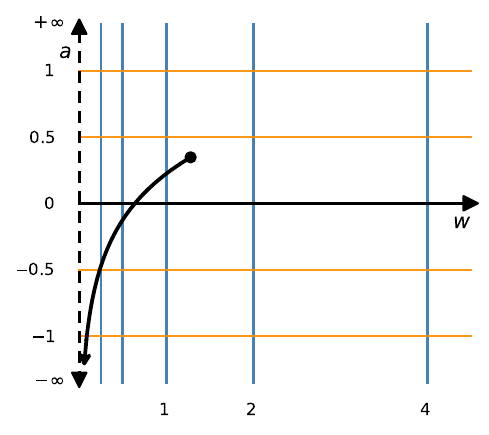}
        \caption{Corresponding symmetry coordinates}
        \label{fig:wa_chart}
    \end{subfigure}
    \caption{In the positive $(u,v)$ quadrant (excluding the axes), the diagonal network $f(x)=uvx$ can be given symmetry coordinates $w=uv$ (blue) and $a=\frac{1}{2}\log(u/v)$ (orange). These are ill-defined at the boundary, as shown by a path (black) approaching the $v$-axis diverging to $(0,-\infty)$ in the $(w,a)$ plane. The finite $a$-axis corresponds to approaching $(u,v)=(0,0)$ along different rays.}
    \label{fig:diag_chart}
\end{figure}
As a simple example of the calculation, we begin with the simplest \emph{diagonal network} \citep{berthier2023DiagLNincrementallearning, pesme2023DiagLNsaddletosaddle, berthier2026DiagLNlasso} $f(x)=uvx$ for a pair of scalar parameters $\theta=(u,v)$ and input $x$. We identify $f$ with the product $w=uv$, noting it has the scaling symmetry $u\mapsto cu, v\mapsto v/c$ for any $c\ne0$ (see \cref{fig:diag_loss} for an example loss landscape). On a fixed branch of the hyperbolic fiber, a convenient choice of symmetry coordinates on the set where $u\ne0$ and $v\ne0$ (which is regular) is $a=\frac12\log|u/v|$, so that
\begin{align*}
u=e^a\sqrt{|w|},     \qquad      v=e^{-a}\text{sign}(w)\sqrt{|w|}
\end{align*}
as depicted in \cref{fig:diag_chart}. The Jacobian and regularization then take the convenient forms
\begin{align*}
d\theta=du\,dv=dw\,da,       \qquad      R(w,a)=\frac{u^2+v^2}{2}=|w|\cosh(2a).
\end{align*}
Ignoring the factor of two from summing over branches, the fiber partition function in this case is
\begin{align*}
Z_\alpha(w)=\int_{-\infty}^\infty e^{-\alpha|w|\cosh(2a)}\,da = K_0(\alpha|w|),
\end{align*}
where $K_0$ is the modified Bessel function of the second kind. While its asymptotic expansions around $\alpha=\infty$ ($T\to0$) and $\alpha=0$ ($\lambda\to0$) are classical \citep{abramowitz1964handbook}, our framework gives simpler paths to eliminating the regularization strength $\lambda$ in $\alpha=\lambda/T$. Namely, taking $T\to0$ first gives effective regularization
\begin{align*}
R_\text{eff}(w) = \inf_{a\in\mathbb R} |w|\cosh(2a) = |w|.
\end{align*}
Suppose instead we fix $T>0$ and take $\lambda\to0$. For this and our subsequent calculations of the scaling of fiber partition functions, the idea is to determine how the regularization cuts off the non-compact symmetry directions. Here, the integrand $\exp(-\alpha|w|\cosh(2a))$ is close to one when $\alpha|w|\cosh(2a)\lesssim 1$, and decays rapidly outside this region. Since $\cosh(2a)$ grows exponentially with $|a|$, the region contributing to $Z_\alpha$ scales like
$|a|\lesssim \frac{1}{2}\log\frac{1}{\alpha} + O_w(1).$
This suggests reparameterizing the fiber coordinate $a=\frac{1}{2}\log\left(\frac{1}{\alpha}\right)s$, such that the relevant region has finite size. Then
\begin{align*}
Z_\alpha(w) = \frac{\log( 1/\alpha)}{2}\int_{-\infty}^\infty e^{-\frac{|w|}{2}\left(\alpha^{1-s}+\alpha^{1+s}\right)}\,ds.
\end{align*}
As $\alpha\to0$, the $\alpha$ terms vanish for $|s|<1$, making the integrand close to one, and at least one diverges for $|s|>1$, making the integrand close to zero. Thus the expanding relevant region of the fiber in $a$-coordinates becomes $s \in [-1,1]$. Letting $C_\alpha = \log (1/\alpha)$, we obtain that
\begin{align*}
\Omega(w)=\lim_{\alpha\to0}\frac{Z_\alpha(w)}{C_\alpha} = \frac{1}{2}\int_{-1}^1\,ds = 1.
\end{align*}
(The limit is justified in \cref{app:diag}.) For this simple example, the entropic term $\log\Omega(w)=0$ is trivial, and the limiting distribution has no symmetry-induced bias: $\lim_{\lambda\to0}p_{\lambda,T}(w)\propto\exp(-\ell(w)/T)$. It is worth pausing to make two observations. First, while we derive a limiting distribution for $w$ as $\lambda\to0$, one cannot do the same in the original $(u,v)$ space, as probability escapes along the non-compact loss valley (see Appendix \cref{fig:diag-uv-dist}). Second, the entropic correction happens to be trivial in this case, but there is no a priori reason for it to be so. In the models we examine later, the renormalized entropy retains a non-trivial dependence on the represented function.

\paragraph{Diagonal networks of arbitrary width and depth} The story for general diagonal networks is much the same. A width $n$ and depth $D$ network is
$f(x)=w^\top x$ where $w_i=\prod_{d=1}^D u_{di}$ with parameters $u_{di}$ for $d\in[D]$ and $i\in[n]$. The regularization $R=\frac{1}{2}\sum_{di} u_{di}^2$ decouples the coordinates $i$. The $T\to0$ effective regularization recovers the known representation cost $R_\text{eff}(w)=\frac{D}{2}\sum_i|w_i|^{2/D}$ \citep{gunasekar2018implicit,zhen2021representationcost}, whereas the $\lambda\to0$ limit gives $\Omega(w)=1$ after renormalizing by $C_\alpha\sim(\log1/\alpha)^{(D-1)n}$.

\begin{figure}[ht]
    \centering
    \includegraphics[width=0.85\textwidth]{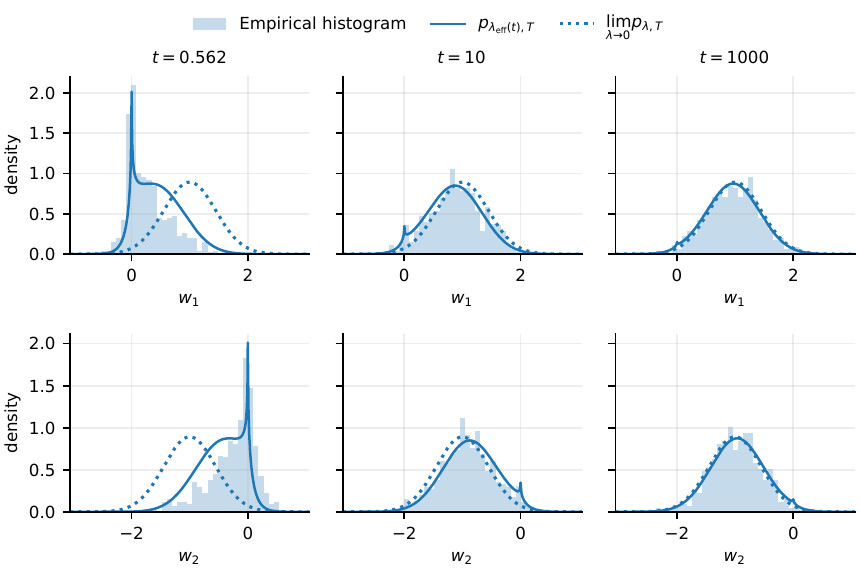}
    \caption{Unregularized Langevin densities for $w_i=u_iv_i$ with $\ell(w)=\|w-w_*\|^2/2$, for $w_*=(1,-1)$ and $T=0.2$, over 512 trials. Finite-time $\lambda=0$ densities are increasingly well predicted by $p_{\lambda_\mathrm{eff}(t),T}(w)$ (blue) at large times, and approach the predicted $\lambda\to0$ limit (blue, dotted). (See Appendix \cref{fig:diag-hist-long} for $t=10^4$.)}
    \label{fig:diag_alpha_eff}
\end{figure}

\paragraph{Connection to unregularized Langevin}
The Bessel function prediction is exact for positive $\lambda$ and $T$, matching simulated Langevin statistics (see Appendix \cref{fig:diag-exact}). We can also ask whether the renormalized entropy is predictive of unregularized dynamics. In particular, it is not clear a priori that we can interchange limits, so that the limiting distribution $\lim_{t\to\infty}p_{t,0,T}(w)$ (if it exists) corresponds to our $\lim_{\lambda\to 0}p_{\lambda,T}(w) = \lim_{\lambda\to 0}\lim_{t\to\infty}p_{t,\lambda,T}(w)$. One idea is to view unregularized dynamics as having a time-dependent effective regularization, with strength inverse to its diffusion away from initialization. This can be made precise by a rescaling of the dynamics, which reveals an instantaneous effective potential $L(\theta)+\lambda_\text{eff}(t)R(\theta)$ with $\lambda_\text{eff}(t) \sim 1/2t$. The dynamics of $\theta$ minimize this increasingly-unregularized potential; but since the potential itself changes over time, the parameters may not have time to equilibrate under it, in particular when they escape to infinity. We show that for diagonal networks, the equilibration is in fact fast enough.
\begin{restatable}{theorem}{diagthm}
\label{thm:diagonal}
For depths $D_i \ge 2$, let $\theta=(u_{di}:i\in[n],d\in [D_i])$ and  $w_i=\prod_{d=1}^{D_i} u_{di}$ parameterize the diagonal network $f(x)=w^\top x$. Suppose $\ell\in C^2(\R^n)$ is convex and coercive, and fix a temperature $T>0$. For regularization $\lambda\ge0$, write $P_{t,\lambda,T}$ for the distribution of $\theta_t$ under Langevin dynamics with potential $\ell(w)+\lambda R(\theta)$, and for $\lambda>0$, let $P_{\lambda,T}$ be the limiting Gibbs distribution. Then for any initial distribution independent of $\lambda$,
\begin{align*}
\| P_{t,\lambda,T} - P_{\lambda_\mathrm{eff}(t,\lambda),T} \|_\mathrm{TV} \to 0,
\qquad
\lambda_\mathrm{eff}(t,\lambda)\sim
\begin{cases}
    \lambda(1-e^{-2\lambda t})^{-1} & \lambda>0\\
    (2t)^{-1} & \lambda=0
\end{cases}
\end{align*}
for any joint limit $t\to\infty$ and $\lambda\to0$ (including $\lambda=0$).
If for small $\lambda$ the (possibly $\lambda$-dependent) initial density has uniformly bounded $L^2$ norm and $\E_{P_{0,\lambda,T}}\|\theta\| + \sqrt{L(\theta)-\min\ell}$ is uniformly bounded,
\begin{align*}
\| P_{t,\lambda,T} - P_{\lambda_\mathrm{eff}(t,\lambda),T} \|_\mathrm{TV} \lesssim \frac{1}{\sqrt{\log( T / \lambda_\mathrm{eff}(t,\lambda))}}.
\end{align*}
\end{restatable}
\begin{proof}
See \cref{app:diag-lim-int}.
\end{proof}
Pushing forward through $\pi$, we get the equivalent result for the distributions over $w$, and under the uniform bounds $p_{t,0,T}(w)$ approaches $\lim_{\lambda\to0}p_{\lambda,T}(w)$ in total variation at rate $O((\log t)^{-1/2})$ (see \cref{app:diag-lim-int}).
The uniform bounds are mild, and hold for example for Gaussian initialization with squared loss.
We conjecture the proof technique generalizes to linear networks, although its applicability to general homogeneous networks is less clear.

We verify the theory experimentally (see \cref{app:diag-experiment} for details). In particular, \cref{fig:diag_alpha_eff} shows the approach of the unregularized empirical distribution to the limit predicted by the simple $\lambda\to0$ calculation, while also being more accurately modeled using $\lambda_\mathrm{eff}(t)$, especially at intermediate times when transient initialization effects decay. (This can be shown quantitatively using the Kolmogorov-Smirnov distance, as in Appendix \cref{fig:diag-cdf,fig:diag-ks}.)

\section{Linear networks}
We encounter our first non-trivial entropic correction in linear networks. Let us begin by considering $f(x)=UV^\top x$ for $x \in \R^n$, and $U,V$ square matrices. We now identify $f$ with $W=UV^\top$, which has the symmetry $U\mapsto UC$ and $V\mapsto VC^{-\top}$ for $C\in\text{GL}(n)$ (an invertible matrix). We work on the regular stratum where $W$ is full-rank, and consider its singular value decomposition $W=PSQ^\top$. Then any $U$ and $V$ may be written as $U=PS^{1/2}C$ and $V=QS^{1/2}C^{-\top}$ for some choice of $C$. The change of measure and regularization are then
\begin{align*}
dU\,dV = |C|^{-n}\,dW\,dC,
\qquad
R= \frac{1}{2}(\|U\|_F^2+\|V\|_F^2) = \frac{1}{2}\Tr\left(S\left(CC^\top+(CC^\top)^{-1}\right)\right),
\end{align*}
where we write $|C|= |\mathrm{det}~C|$. The fiber partition function has a closed form in terms of a matrix Bessel function. Of interest to us, however, are the two limiting regimes described previously.

Taking $T\to0$, and noting $CC^\top$ is positive definite, gives effective nuclear norm regularization
\begin{align*}
R_\text{eff}(W)
=\frac{1}{2}\inf_{X\in\text{PD}(n)}\Tr\left(S\left(X+X^{-1}\right)\right) = \Tr(S) = \|W\|_*
\end{align*}
the second-to-last equality being a lower bound which is tight when $X$ is the identity. This corresponds to the matrix factorization results of \citet{srebro2004maxmargin,zhen2021representationcost}. For the $\lambda\to0$ limit, we examine the fiber partition function
\begin{align*}
Z_\alpha(W) = \int e^{-\frac{\alpha}{2}\Tr\left(S \left(C C^\top +  \left(C C^\top\right)^{-1} \right)\right)} d\mu(C)
\propto
\int_{X\succ 0} e^{-\frac{\alpha}{2}\Tr\left(S \left(X+  X^{-1} \right)\right)}|X|^{-(n+1)/2} \,dX.
\end{align*}
A generalization of the argument from \cref{sec:diag} (\cref{app:linear-entropy}), centered around the integrand being non-negligible where $X$ has eigenvalues $\alpha\lesssim \lambda_i(X) \lesssim \frac{1}{\alpha}$, gives that $Z_\alpha(W)$ scales as
\begin{align*}
\alpha^{-kr}\left(\log\frac{1}{\alpha^2}\right)^{\mathbf{1}\{n\text{ is odd}\}}\int_{\text{Gr}(k,n)} \left|Q^\top S^2 Q\right|^{-r/2}\,d\nu_{k,n}(Q).
\end{align*}
Here, $k=\floor{n/2}$, and $r=n-k=\ceil{n/2}$, which maximizes $\alpha^{-kr}$  over the choices of $k$ for $\alpha\to0$.
The measure $\nu_{k,n}$ is the invariant probability measure on the Grassmannian, the space of $k$-dimensional subspaces of $\R^n$, and $Q$ is any orthonormal basis of the chosen subspace. The renormalized entropy is thus
\begin{align*}
\log \Omega(W) = \log \int_{\text{Gr}(k,n)} \text{Vol}_k(W;Q)^{-r} \,d\nu_{k,n}(Q),
\end{align*}
where $\text{Vol}_k(W;Q)=|Q^\top W^\top W Q|^{1/2}$ measures how much $W$ expands $k$-dimensional volume in the subspace spanned by $Q$. (Note we can replace $S^2$ by $W^\top W$ in the integrand by basis-invariance). The entropic bias thus has an intuitive interpretation: those $W$ are preferred which, on average, contract $k$-dimensional subspaces. By the variational characterization of the singular values $s_1\le\cdots\le s_n$,
we have $\min_{Q} \text{Vol}_k(W;Q)  = s_1\cdots s_k$,
so the bias is towards matrices whose products of the $k$ smallest singular values are close to zero.

\begin{figure}[t]
    \centering
    \includegraphics[width=0.85\textwidth]{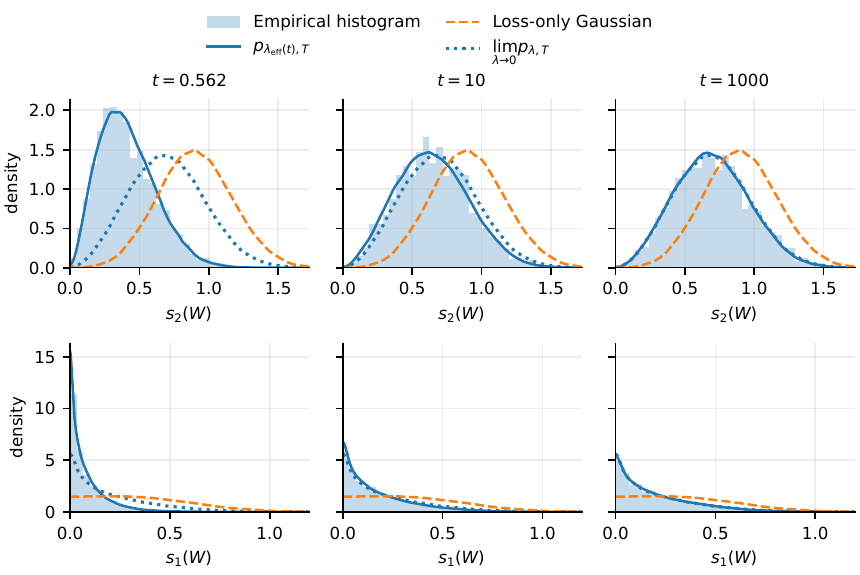}
    \caption{Langevin densities of the two smallest singular values of $W=UV^\top$ with $\ell(W)=\|W-W_*\|_F^2/2$ for $W_*=I_4$ and $T=0.2$, over 2048 trials. Empirical $\lambda=0$ densities appear to track the $\lambda_\text{eff}(t)$ prediction (blue), and approach $\lim_{\lambda\to0}p_{\lambda,T}$ (blue, dotted), as opposed to the na\"ive distribution without the entropic correction (orange, dashed). (See Appendix \cref{fig:linear-histogram-long} for all four singular values up to $t=10^4$.)}
    \label{fig:linear_alpha_eff}
\end{figure}

\paragraph{Comparison to unregularized Langevin} As with diagonal networks, we can ask whether the entropic correction above is predictive of unregularized Langevin dynamics. We test this empirically (\cref{fig:linear_alpha_eff,app:linear-experiment}). From the theory, the smaller singular values skew towards zero in the $\lambda\to0$ limit. The long-time empirical distribution appears to agree, while at intermediate times it agrees closely with the Gibbs distribution under $\lambda_\mathrm{eff}(t)$, which leads us to conjecture \cref{thm:diagonal} may generalize. (We make these comparisons quantitative in Appendix \cref{fig:linear-cdf,fig:linear-ks}).

\section{Homogeneous neural networks}
We turn to neural networks $f(x)=W_D\sigma(W_{D-1}\cdots\sigma(W_1x))$ with $W_d \in \mathbb R^{n_d \times n_{d-1}}$ and one-homogeneous activation, $\sigma(cz)=c\sigma(z)$ for $c>0$. This includes ReLU and leaky ReLU. (For homogeneity of different orders see \cref{app:two-layer-homog}.) Such networks have neuron-wise scaling symmetry $W_d \mapsto P_d W_d P_{d-1}^{-1}$, where $P_d$ are positive diagonal matrices with $P_0$ and $P_D$ identity matrices. We introduce coordinates by recursively normalizing each layer's incoming weights: $W_d=P_d \bar W_d P_{d-1}^{-1}$ where $\| (\bar W_d)_{i\cdot}\|=1$ for all $d<D$ and $i\in[n_d]$, and $P_d=\text{diag}(\rho_d)$. The change of measure and regularization have simple forms in terms of edges $e=(u,v)$ between nodes:
\begin{align*}
dW=d\omega(\bar W_{1:D-1})\,d\bar W_D \prod_{\text{hidden } v} \rho_v^{\delta_v-1}\,d\rho_v
\qquad
R(\bar W,\rho)=\frac{1}{2}\sum_{e=(u,v)}\bar W_e^2\left(\frac{\rho_v}{\rho_u}\right)^2
\end{align*}
where $d\omega$ is the product of spherical surface measures on the rows of $\bar W_d$ for $d<D$, and $\delta_v$ is the in-degree of $v$ minus its out-degree. The normalized weights $\bar W$ are reduced parameters eliminating the node scaling symmetry, but may still contain ``hidden symmetries'' \citep{rolnick2020reverse,grigsby2023hidden}, whose general characterization is open \citep{grillo2026identifiable}. On regular regions where scaling is the only local symmetry, $f$ locally determines $\bar W$ and $(\bar W,P)$ are the symmetry coordinates of \cref{sec:theory}. Otherwise, integrating over $P$ gives implicit bias just over $\bar W$.

With that in mind, the $T\to0$ limit gives, with an abuse of notation
\begin{align*}
R_\text{eff}(f)=\inf_{\bar W:f_{\bar{W}}=f} R_\text{eff}(\bar{W}),
\qquad
R_\text{eff}(\bar W)=\inf_{\rho>0} R(\bar W,\rho).
\end{align*}
Differentiating $R(\bar W, \rho)$ with respect to $\log \rho_v$ for any node $v$, we find the infimum in $R_\text{eff}(\bar W)$ is attained when $\sum_{(u,v)}  W_{uv}^2 = \sum_{(v,z)} W_{vz}^2$, at \emph{balance} \citep{stock2019equinormalization}. Minimizing over any remaining hidden symmetries in $R_\text{eff}(f)$ recovers a version of the cost of \citet{neyshabur2015norm}.

For $\lambda\to0$ we consider a tractable but illustrative case: all layers having the same width $n_d=n$. Then $\delta_v=0$ for all nodes $v$. Following the diagonal network case and letting $\rho_v=\alpha^{-s_v/2}$,
\begin{align*}
Z_\alpha(\bar W)=\left(\frac12\log\frac{1}{\alpha}\right)^{(D-1)n} \int \exp\left(-\frac{1}{2}\sum_{e=(u,v)}\bar W_e^2\alpha^{1-(s_v-s_u)}\right)\,\prod_vds_v.
\end{align*}
As $\alpha\to0$ the integrand approaches one when $s_v-s_u<1$ for all $e=(u,v)$, and vanishes otherwise. Exchanging limits (as justified in \cref{app:square-relu}), we get $Z_\alpha(\bar W) C_{D,n}\sim \log(1/\alpha)^{(D-1)n}$. We make two observations. First, there is no entropic selection among choices of $\bar W$ on this stratum, but as we exhibit in the next section, this does not preclude entropic bias in the function $f$ itself. (Different layer shapes do give non-trivial entropy for $\bar W$ encouraging feature sparsity. See \cref{app:square-to-scalar-relu}.) Second, as noted above, integrating over $P$ may not integrate over all of $\pi^{-1}(f)$. Despite this, we can determine the complete implicit bias in $f$ in the case of ReLU networks without characterizing the hidden symmetries. A direct calculation shows the zero function dominates the entropy.

\begin{theorem}
Let $\pi:\theta\mapsto f$ be homogeneous, and suppose $\ell$ is bounded below and radially coercive, meaning $\ell(cf)\to\infty$ as $c\to\infty$ for any non-zero $f\in\pi(\Theta)$. Write $\Pr_{\lambda,T}$ for the Gibbs distribution under $\ell(f)+\lambda R(\theta)$. If $\mathcal D=\{\theta:\pi(\theta)=0\}$ has positive mass, then $\Pr_{\lambda,T}(f=0)\to1$ as $\lambda\to0$.
\end{theorem}
\begin{proof}
Since $\pi(c\theta)=c^D\pi(\theta)$, the dead region $\mathcal{D}$ is closed under positive scaling. Therefore, by the change of variables $\theta=\alpha^{-1/2}\xi$,
\begin{align*}
\nu_\alpha(\{0\}) = \int_\mathcal{D}e^{-\alpha\|\theta\|^2/2}\,d\theta = (2\pi)^{N/2}\alpha^{-N/2}\Pr(\xi\in \mathcal{D})
\end{align*}
for $\xi\sim\mathcal N(0,I_N)$. Using the same change of variables in the denominator,
\begin{align*}
\Pr\nolimits_{\lambda,T}(f=0) = \frac{e^{-\ell(0)/T}\nu_\alpha(\{0\})}{\int e^{-\ell(f)/T}\nu_\alpha(df)} = \frac{(2\pi/\alpha)^{N/2}e^{-\ell(0)/T}\Pr(\xi\in\mathcal D)} {(2\pi/\alpha)^{N/2} \mathbb E\left[e^{-\ell(\alpha^{-D/2}\pi(\xi))/T}\right]}.
\end{align*}
Suppose the loss $\ell$ is radially coercive, meaning $\ell(cf)\to\infty$ as $c\to\infty$ for any non-zero $f \in \pi(\Theta)$. Then the term inside the expectation converges to $\exp(-\ell(0)/T)\mathbf 1_\mathcal{D}$. The dominated convergence theorem implies the denominator approaches the numerator as $\alpha\to0$.
\end{proof}

In bias-free ReLU networks of at least two hidden layers (as well as networks with hidden biases), the positive mass assumption holds because $\mathcal{D}$ contains an open set, a layer of negative weights zeroing out later activations. The radial coercivity assumptions holds, for example, for population squared error. (For empirical squared error, one in general gets convergence to zero \emph{on the training data}, by identifying $f$ with its evaluation thereon.)
The proof then shows that the dead stratum's volume growth $\alpha^{-N/2}$ is much faster than the logarithmic growth on the regular region considered previously.
This means that despite potentially having a much higher loss than other functions, the zero function dominates in the $\lambda\to0$ limit. We call this concentration on $f=0$ \emph{neural heat death}.

\begin{figure}[t]
    \centering
    \includegraphics[width=\textwidth]{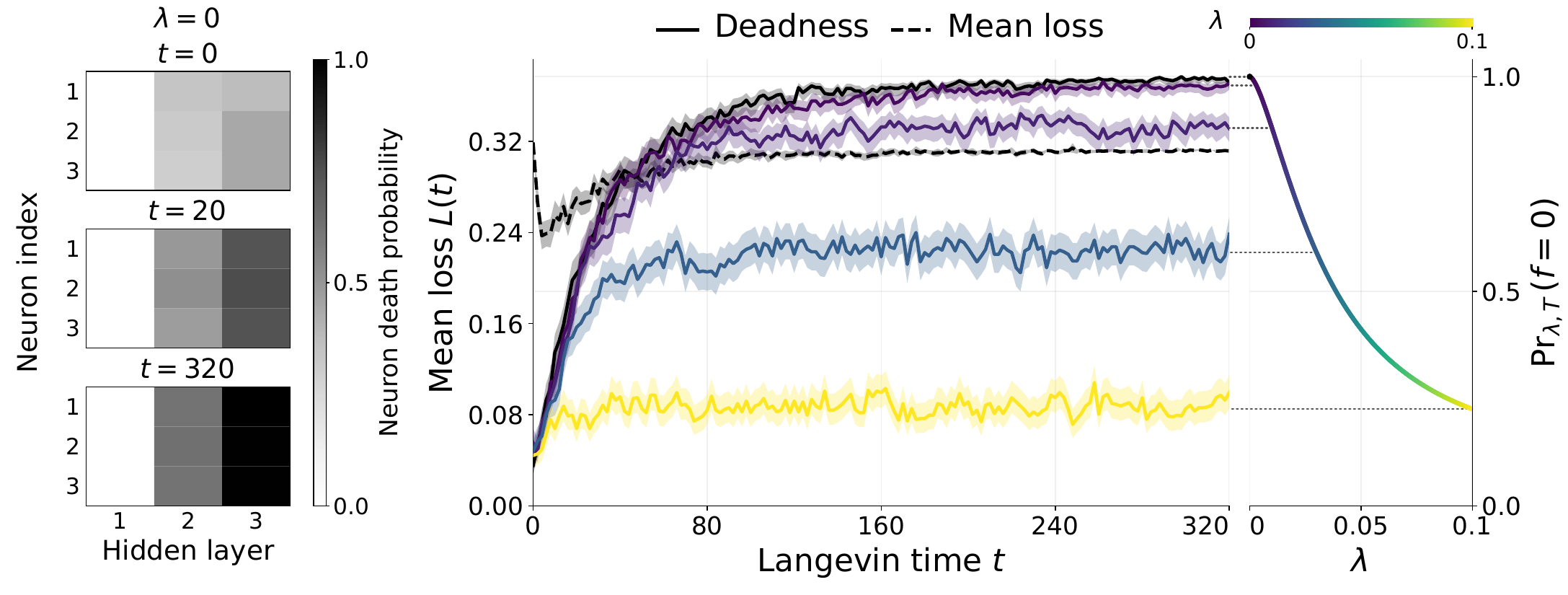}
    \caption{Langevin dynamics for a 3-hidden-layer ReLU network under squared loss on input data $x=\pm1$ and target $f_*(x)=\operatorname{ReLU}(x)+\tfrac12\operatorname{ReLU}(-x)$, with $T=0.2$ and $\lambda\in\{0,0.003,0.01,0.03,0.1\}$, with 512 trials. For $\lambda=0$, the probability a neuron is zero for all inputs increases over time (heatmaps); the probability that the function outputs zero thus goes to one while loss increases (middle plot, solid and dashed black lines). For $\lambda>0$ (shown in color), the final empirical deadness (middle plot) agrees with the theoretical Gibbs equilibrium (right plot and dotted lines). Shading shows the empirical 95\% interval.}
    \label{fig:relu_heat_death}
\end{figure}

We check empirically whether our prediction agrees with unregularized Langevin dynamics (\cref{app:heat-death-experiment}). \cref{fig:relu_heat_death} shows the progressive heat death of an unregularized ReLU network over time, with a smooth transition in the equilibrium as $\lambda\to0$. (See Appendix \cref{fig:relu_ks} for a comparison of the Kolmogorov-Smirnov distance to the na\"ive prediction.)

\section{Simplicity bias and the Neural Tangent Kernel}

In multiple examples --- linear networks, heat death, as well as non-constant width homogeneous networks (see \cref{app:square-to-scalar-relu}) --- we see an entropic bias towards some form of sparsity. We have also seen multiple examples of models where the entropy correction in effective parameters appears trivial, such as on a regular stratum of square ReLU networks. This, however, does not mean there is no symmetry-induced implicit bias in $f$. In fact, as we show below, \emph{even when} the effective parameters are (locally) identifiable there is a bias towards sparsity, at the level of entropy of functions $f$. The key is the change in volume captured by the NTK.

\paragraph{Simplicity bias via the NTK} Suppose we are on a regular stratum on which $\text{rank}(D\pi_\theta) = k$. To define a density on functions, we fix an input distribution $x\sim Q$ and take $df=d\mathcal H^k_Q(f)$, the induced Hausdorff measure (\cref{app:functional-entropy}). By the coarea formula (see e.g.\ \citet{simon2018gmt}), the fiber partition function with respect to $df$ is
\begin{align*}
Z_\alpha(f) = \int_{\pi^{-1}(f)}\frac{e^{-\alpha R(\theta)}}{\sqrt{\det^+K(\theta)}}d\mathcal{H}^{N-k}(\theta),
\end{align*}
where $K(\theta)=D\pi_\theta D\pi_\theta^\top$ is the NTK and $\det^+$ denotes the product of non-zero eigenvalues. From this we see that functions are preferred when they vary slowly with the parameters. This is intuitive: Langevin dynamics spends more time near functions that are insensitive to parameter motion. (Note that this preference holds on each stratum, rather than describing selection \emph{between} strata.)

The same interpretation applies after symmetry coordinates have been eliminated and replaced with identifiable effective parameters $\bar\theta$, where $\pi(\theta)=\bar\pi(\bar\theta)$, to get $\Omega(\bar\theta)$. Then the integral above is replaced by the single point $\bar\theta=\bar\pi^{-1}(f)$. Replacing $K$ by $\bar K$ (defined using $\bar\pi$ in the obvious way), and taking $\alpha\to0$ we get the entropic correction $\log\Omega(f)=\log\Omega(\bar\theta)-\frac{1}{2}\log\det^+(\bar K(\bar\theta))$.

\paragraph{A simple example} Consider two-layer ReLU networks $f(x)=\sum_{i=1}^n u_i\sigma(v_i^\top x)$ where $x\in\R^{n_0}$. Following the normalization procedure from the previous section, we get $f(x)=\sum_{i=1}^n r_i\sigma(\varphi_i^\top x)$ where the $\varphi_i$ are unit vectors and $r_i = \|v_i\|u_i$. By our usual recipe (\cref{app:two-layer-homog}), we find $\log \Omega(r,\varphi)=0$ relative to $dr\,d\omega(\varphi)$. Note that if all $r_i\ne0$ and $\varphi_i\ne\pm\varphi_j$ for $i\ne j$, then $r$ and the hyperplanes normal to $\varphi$ are determined by $f$ up to permutation \citep{dereich2022minimal}, making these effective parameters for data distributions with positive density on a neighborhood of the origin. Following a brief calculation of the neural tangent features (under minor assumptions; \cref{app:functional-entropy}), we find that outside of zero or colliding neurons
\begin{align*}
\log \Omega(f) = -(n_0-1)\sum_{i=1}^n\log|r_i| + \text{$\varphi$ term}.
\end{align*}
From this we can directly read off a preference towards neurons of small scale $r_i$. The mechanism is exactly a soft NTK sparsity preference: as $r_i$ decreases, changes in $\varphi_i$ have less effect on $f$. 
This is akin to the fact that sampling polar coordinates uniformly does not uniformly sample points in a disk: samples are more dense near the origin, since changes in angle have less of an effect there.

\section{Discussion and related work}

Our Langevin framework naturally affords and relates two limiting regimes. Taking $T\to 0$ minimizes regularization over the fiber first, giving an effective regularization. This recovers representation cost theories \citep{zhen2021representationcost}. While these have been shown not to describe gradient flow in general \citep{arora2019implicitregularizationdeepmatrix}, we find them relevant to the $T\to0$ limit of Langevin dynamics. The balance condition we recover for deep homogeneous networks occurs across gradient flow analyses, in the form of conservation laws \citep{saxe2014exactsolutionsnonlineardynamics,du2018algorithmic,kunin2021neuralmechanicssymmetrybroken}. On the other hand, works on SGD (with positive temperature) find generalizations of balance \citep{ziyin2024noiseequilibrium,ziyin2025thermodynamics} or other dynamically attractive conditions \citep{chen2023stochasticcollapse}. This attraction within fibers highlights a difference between SGD and our other limiting regime.

Taking $\lambda\to0$ with $T>0$ instead reveals entropic selection based on the asymptotic growth of fibers. \citet{aladrah2026geometric} develop a closely related perspective on symmetry-induced bias, but handle non-compact symmetries by gauge-fixing. This essentially assumes the balanced case, consistently with their experiments, which use SGD. Similarly, \citet{menon2025entropy} derive an entropy on the balanced manifold of deep linear networks, whose symmetry is compact. 
In contrast to these works, we do not assume balance, studying the entire non-compact fiber (though we do recover the \citet{menon2025entropy} entropy as a subleading term in the $T\to0$ expansion of $Z_\alpha$ in \cref{app:menon-yu-entropy}).
%Our results complement empirical studies of simplicity bias which suggest parameter volume as a mechanism \citep{valleperez2019simple}, and our explicit linear network entropy may help explain observed low-rank bias \citep{huh2023lowrank}.

Analysis of neural network priors when $\lambda>0$ is established \citep{li2021renormalization,hanin2023interpolation}. Particularly relevant are \citet{zavatoneveth2021exact,noci2021prior}, who find exact finite-width priors for linear and ReLU networks using Meijer-G functions, which include the Bessel functions we find for diagonal networks as a special case. In particular, \citet{zavatoneveth2021exact} find an atom at zero output for a fixed input. Heat death extends this result by showing the Gibbs probability of outputting zero on $\emph{all}$ inputs goes to one as the prior variance increases. Related work has shown ReLU networks at initialization can die as depth increases, at finite parameter variance \citep{lu2020dying,luther2020reexamining}. Our work also bears some resemblance to singular learning theory \citep{watanabe2013wbic}. In particular, \citet{aoyagi2005reducedrank} give the scaling the \emph{total} partition function under squared loss with zero target; our calculation resolves the density factor $\Omega(W)$, independently of the loss.

Our work opens the door to interesting questions we do not answer here, such as what the renormalized entropy is for other architectures, and when the limit interchange between weak regularization and unregularized Langevin is formally justified beyond the diagonal case. Additionally, what can be said under other kinds of noise for which a Gibbs equilibrium may not exist, such as SGD? We leave these questions to future work.

\paragraph{Acknowledgments} All authors were supported by the Gatsby Charitable Foundation. This work also was supported by a Research Grant from HFSP (Ref.-No: RGEC33/2024) to A.S., CIFAR (Learning in Machines and Brains), and the Sainsbury Wellcome Centre Core Grant from Wellcome (219627/Z/19/Z).

\bibliography{iclr2027_conference}

@inproceedings{pesme2023DiagLNsaddletosaddle,
      title={Saddle-to-Saddle Dynamics in Diagonal Linear Networks}, 
      author={Scott Pesme and Nicolas Flammarion},
      year={2023},
      booktitle={Advances in Neural Information Processing Systems},
      volume={36},
      pages={7475--7505},
      doi={10.52202/075280-0329},
      eprint={2304.00488},
      archivePrefix={arXiv},
      primaryClass={cs.LG},
      url={https://papers.neurips.cc/paper_files/paper/2023/hash/17a9ab4190289f0e1504bbb98d1d111a-Abstract-Conference.html}, 
}

@misc{berthier2026DiagLNlasso,
      title={Diagonal Linear Networks and the Lasso Regularization Path}, 
      author={Raphaël Berthier},
      year={2026},
      note={Version 3, 18 March 2026. First submitted 23 September 2025},
      eprint={2509.18766v3},
      archivePrefix={arXiv},
      primaryClass={cs.LG},
      url={https://arxiv.org/abs/2509.18766v3}, 
}

@inproceedings{arora2019implicitregularizationdeepmatrix,
      title={Implicit Regularization in Deep Matrix Factorization}, 
      author={Sanjeev Arora and Nadav Cohen and Wei Hu and Yuping Luo},
      year={2019},
      booktitle={Advances in Neural Information Processing Systems},
      volume={32},
      eprint={1905.13655},
      archivePrefix={arXiv},
      primaryClass={cs.LG},
      url={https://proceedings.neurips.cc/paper/2019/hash/c0c783b5fc0d7d808f1d14a6e9c8280d-Abstract.html}, 
}

@inproceedings{zhen2021representationcost,
 author = {Dai, Zhen and Karzand, Mina and Srebro, Nathan},
 booktitle = {Advances in Neural Information Processing Systems},
 editor = {M. Ranzato and A. Beygelzimer and Y. Dauphin and P.S. Liang and J. Wortman Vaughan},
 pages = {26884--26896},
 publisher = {Curran Associates, Inc.},
 title = {Representation Costs of Linear Neural Networks: Analysis and Design},
 url = {https://proceedings.neurips.cc/paper_files/paper/2021/file/e22cb9d6bbb4c290a94e4fff4d68a831-Paper.pdf},
 volume = {34},
 year = {2021}
}

@article{berthier2023DiagLNincrementallearning,
      title={Incremental Learning in Diagonal Linear Networks}, 
      author={Raphaël Berthier},
      year={2023},
      journal={Journal of Machine Learning Research},
      volume={24},
      number={171},
      pages={1--26},
      eprint={2208.14673},
      archivePrefix={arXiv},
      primaryClass={cs.LG},
      url={https://jmlr.org/papers/v24/22-1395.html}, 
}

@article{zhao2025symmetryneuralnetworkparameter,
      title={Symmetry in Neural Network Parameter Spaces}, 
      author={Bo Zhao and Robin Walters and Rose Yu},
      year={2025},
      journal={Transactions on Machine Learning Research},
      eprint={2506.13018},
      archivePrefix={arXiv},
      primaryClass={cs.LG},
      url={https://openreview.net/forum?id=jLpWq5QY6I}, 
}

@inproceedings{kunin2021neuralmechanicssymmetrybroken,
      title={Neural Mechanics: Symmetry and Broken Conservation Laws in Deep Learning Dynamics}, 
      author={Daniel Kunin and Javier Sagastuy-Brena and Surya Ganguli and Daniel L. K. Yamins and Hidenori Tanaka},
      year={2021},
      booktitle={International Conference on Learning Representations},
      eprint={2012.04728},
      archivePrefix={arXiv},
      primaryClass={cs.LG},
      url={https://openreview.net/forum?id=q8qLAbQBupm}, 
}

@inproceedings{saxe2014exactsolutionsnonlineardynamics,
      title={Exact solutions to the nonlinear dynamics of learning in deep linear neural networks}, 
      author={Andrew M. Saxe and James L. McClelland and Surya Ganguli},
      year={2014},
      booktitle={International Conference on Learning Representations},
      eprint={1312.6120},
      archivePrefix={arXiv},
      primaryClass={cs.NE},
      url={https://arxiv.org/abs/1312.6120}, 
}

@inproceedings{jacot2018ntk,
  author = {Arthur Jacot and Franck Gabriel and Cl{\'e}ment Hongler},
  title = {{Neural Tangent Kernel: Convergence and Generalization in Neural Networks}},
  year = {2018},
  url = {https://arxiv.org/abs/1806.07572},
  booktitle = {Advances in Neural Information Processing Systems},
  volume = {31},
  eprint = {1806.07572},
  archivePrefix = {arXiv},
}

@inproceedings{marcotte2025intrinsic,
  author = {Sibylle Marcotte and Gabriel Peyr{\'e} and R{\'e}mi Gribonval},
  title = {{Intrinsic training dynamics of deep neural networks}},
  year = {2026},
  booktitle = {International Conference on Learning Representations},
  url = {https://proceedings.iclr.cc/paper_files/paper/2026/hash/184b2454410a88c280d29aa95fbac6e2-Abstract-Conference.html},
  eprint = {2508.07370},
  archivePrefix = {arXiv},
}

@inproceedings{bordelon2022selfconsistent,
  author = {Blake Bordelon and Cengiz Pehlevan},
  title = {{Self-Consistent Dynamical Field Theory of Kernel Evolution in Wide Neural Networks}},
  year = {2022},
  url = {https://arxiv.org/abs/2205.09653},
  booktitle = {Advances in Neural Information Processing Systems},
  volume = {35},
  eprint = {2205.09653},
  archivePrefix = {arXiv},
}

@article{mandt2017sgd,
  author = {Stephan Mandt and Matthew D. Hoffman and David M. Blei},
  title = {{Stochastic Gradient Descent as Approximate Bayesian Inference}},
  year = {2017},
  url = {https://jmlr.org/papers/v18/17-214.html},
  journal = {Journal of Machine Learning Research},
  volume = {18},
  number = {134},
  pages = {1--35},
}

@inproceedings{chaudhari2018variational,
  author = {Pratik Chaudhari and Stefano Soatto},
  title = {{Stochastic gradient descent performs variational inference, converges to limit cycles for deep networks}},
  year = {2018},
  url = {https://arxiv.org/abs/1710.11029},
  booktitle = {International Conference on Learning Representations},
  eprint = {1710.11029},
  archivePrefix = {arXiv},
}

@inproceedings{yaida2019fluctuation,
  author = {Sho Yaida},
  title = {{Fluctuation-dissipation relations for stochastic gradient descent}},
  year = {2019},
  url = {https://arxiv.org/abs/1810.00004},
  booktitle = {International Conference on Learning Representations},
  eprint = {1810.00004},
  archivePrefix = {arXiv},
}

@inproceedings{zhang2022invariant,
  author = {Jingzhao Zhang and Haochuan Li and Suvrit Sra and Ali Jadbabaie},
  title = {{Neural Network Weights Do Not Converge to Stationary Points: An Invariant Measure Perspective}},
  year = {2022},
  url = {https://proceedings.mlr.press/v162/zhang22q.html},
  booktitle = {Proceedings of the International Conference on Machine Learning},
  series = {Proceedings of Machine Learning Research},
  volume = {162},
  pages = {26330--26346},
}

@inproceedings{welling2011sgld,
  author = {Max Welling and Yee Whye Teh},
  title = {{Bayesian Learning via Stochastic Gradient Langevin Dynamics}},
  year = {2011},
  url = {https://www.stats.ox.ac.uk/~teh/research/compstats/WelTeh2011a.pdf},
  booktitle = {Proceedings of the International Conference on Machine Learning},
}

@article{kappel2015plasticity,
  author = {David Kappel and Stefan Habenschuss and Robert Legenstein and Wolfgang Maass},
  title = {{Network Plasticity as Bayesian Inference}},
  year = {2015},
  url = {https://journals.plos.org/ploscompbiol/article?id=10.1371/journal.pcbi.1004485},
  journal = {PLOS Computational Biology},
  volume = {11},
  number = {11},
  pages = {e1004485},
  doi = {10.1371/journal.pcbi.1004485},
}

@article{aitchison2021plasticity,
  author = {Laurence Aitchison and Jannes Jegminat and Jorge Aurelio Menendez and Jean-Pascal Pfister and Alexandre Pouget and Peter E. Latham},
  title = {{Synaptic plasticity as Bayesian inference}},
  year = {2021},
  url = {https://www.nature.com/articles/s41593-021-00809-5},
  journal = {Nature Neuroscience},
  volume = {24},
  pages = {565--571},
  doi = {10.1038/s41593-021-00809-5},
}

@inproceedings{zavatoneveth2021exact,
  author = {Jacob A. Zavatone-Veth and Cengiz Pehlevan},
  title = {{Exact marginal prior distributions of finite Bayesian neural networks}},
  year = {2021},
  url = {https://arxiv.org/abs/2104.11734},
  booktitle = {Advances in Neural Information Processing Systems},
  volume = {34},
  pages = {3364--3375},
  eprint = {2104.11734},
  archivePrefix = {arXiv},
}

@inproceedings{noci2021prior,
  author = {Lorenzo Noci and Gregor Bachmann and Kevin Roth and Sebastian Nowozin and Thomas Hofmann},
  title = {{Precise characterization of the prior predictive distribution of deep ReLU networks}},
  year = {2021},
  url = {https://proceedings.neurips.cc/paper_files/paper/2021/hash/ae4503ec3da32f5e9033604744ec45ae-Abstract.html},
  booktitle = {Advances in Neural Information Processing Systems},
  volume = {34},
}

@inproceedings{srebro2004maxmargin,
  author = {Nathan Srebro and Jason D. M. Rennie and Tommi S. Jaakkola},
  title = {{Maximum-Margin Matrix Factorization}},
  year = {2004},
  url = {https://proceedings.neurips.cc/paper/2004/hash/e0688d13958a19e087e123148555e4b4-Abstract.html},
  booktitle = {Advances in Neural Information Processing Systems},
  volume = {17},
}

@inproceedings{gunasekar2017matrix,
  author = {Suriya Gunasekar and Blake Woodworth and Srinadh Bhojanapalli and Behnam Neyshabur and Nathan Srebro},
  title = {{Implicit Regularization in Matrix Factorization}},
  year = {2017},
  url = {https://arxiv.org/abs/1705.09280},
  booktitle = {Advances in Neural Information Processing Systems},
  volume = {30},
  eprint = {1705.09280},
  archivePrefix = {arXiv},
}

@inproceedings{neyshabur2015norm,
  author = {Behnam Neyshabur and Ryota Tomioka and Nathan Srebro},
  title = {{Norm-Based Capacity Control in Neural Networks}},
  year = {2015},
  url = {https://proceedings.mlr.press/v40/Neyshabur15.html},
  booktitle = {Proceedings of the Conference on Learning Theory},
  series = {Proceedings of Machine Learning Research},
  volume = {40},
  pages = {1376--1401},
}

@inproceedings{stock2019equinormalization,
  author = {Pierre Stock and Benjamin Graham and R{\'e}mi Gribonval and Herv{\'e} J{\'e}gou},
  title = {{Equi-normalization of Neural Networks}},
  year = {2019},
  url = {https://arxiv.org/abs/1902.10416},
  booktitle = {International Conference on Learning Representations},
  eprint = {1902.10416},
  archivePrefix = {arXiv},
}

@inproceedings{grigsby2023hidden,
  author = {J. Elisenda Grigsby and Kathryn Lindsey and David Rolnick},
  title = {{Hidden Symmetries of ReLU Networks}},
  year = {2023},
  url = {https://proceedings.mlr.press/v202/grigsby23a.html},
  booktitle = {Proceedings of the International Conference on Machine Learning},
  series = {Proceedings of Machine Learning Research},
  volume = {202},
  pages = {11734--11760},
}

@inproceedings{rolnick2020reverse,
  author = {David Rolnick and Konrad Kording},
  title = {{Reverse-engineering deep ReLU networks}},
  year = {2020},
  url = {https://proceedings.mlr.press/v119/rolnick20a.html},
  booktitle = {Proceedings of the International Conference on Machine Learning},
  series = {Proceedings of Machine Learning Research},
  volume = {119},
  pages = {8178--8187},
}

@misc{grillo2026identifiable,
  author = {Moritz Grillo and Guido Mont{\'u}far},
  title = {{Most ReLU Networks Admit Identifiable Parameters}},
  year = {2026},
  url = {https://arxiv.org/abs/2605.03601},
  eprint = {2605.03601},
  archivePrefix = {arXiv},
  howpublished = {arXiv preprint},
}

@article{dereich2022minimal,
  author = {Steffen Dereich and Sebastian Kassing},
  title = {{On minimal representations of shallow ReLU networks}},
  year = {2022},
  url = {https://arxiv.org/abs/2108.05643},
  journal = {Neural Networks},
  volume = {148},
  pages = {121--128},
  doi = {10.1016/j.neunet.2022.01.006},
  eprint = {2108.05643},
  archivePrefix = {arXiv},
}

@book{simon2018gmt,
  author = {Leon Simon},
  title = {{Introduction to Geometric Measure Theory}},
  year = {2018},
  url = {https://math.stanford.edu/~lms/ntu-gmt-text.pdf},
  publisher = {Stanford University},
  note = {Lecture notes, 2017--2018 version; Chapter 2, Section 7, Theorem 7.3 and Remark 7.4(1), p. 76},
}

@article{li2021renormalization,
  author = {Qianyi Li and Haim Sompolinsky},
  title = {{Statistical Mechanics of Deep Linear Neural Networks: The Back-Propagating Kernel Renormalization}},
  year = {2021},
  url = {https://arxiv.org/abs/2012.04030},
  journal = {Physical Review X},
  volume = {11},
  pages = {031059},
  doi = {10.1103/PhysRevX.11.031059},
  eprint = {2012.04030},
  archivePrefix = {arXiv},
}

@inproceedings{izmailov2021posteriors,
  author = {Pavel Izmailov and Sharad Vikram and Matthew D. Hoffman and Andrew Gordon Wilson},
  title = {{What Are Bayesian Neural Network Posteriors Really Like?}},
  year = {2021},
  url = {https://proceedings.mlr.press/v139/izmailov21a.html},
  booktitle = {Proceedings of the International Conference on Machine Learning},
  series = {Proceedings of Machine Learning Research},
  volume = {139},
  pages = {4629--4640},
}

@article{hanin2023interpolation,
  author = {Boris Hanin and Alexander Zlokapa},
  title = {{Bayesian Interpolation with Deep Linear Networks}},
  year = {2023},
  url = {https://arxiv.org/abs/2212.14457},
  journal = {Proceedings of the National Academy of Sciences},
  volume = {120},
  number = {23},
  pages = {e2301345120},
  doi = {10.1073/pnas.2301345120},
  eprint = {2212.14457},
  archivePrefix = {arXiv},
}

@inproceedings{kobialka2026priors,
  author = {Julius Kobialka and Emanuel Sommer and Chris Kolb and Juntae Kwon and Daniel Dold and David R{\"u}gamer},
  title = {{On the Interplay of Priors and Overparametrization in Bayesian Neural Network Posteriors}},
  year = {2026},
  series = {Proceedings of Machine Learning Research},
  volume = {300},
  pages = {19--27},
  url = {https://proceedings.mlr.press/v300/kobialka26a.html},
  booktitle = {Proceedings of the 29th International Conference on Artificial Intelligence and Statistics},
  eprint = {2603.22030},
  archivePrefix = {arXiv},
}

@article{menon2025entropy,
  author = {Govind Menon and Tianmin Yu},
  title = {{An Entropy Formula for the Deep Linear Network}},
  year = {2026},
  journal = {SIAM Journal on Mathematical Analysis},
  volume = {58},
  number = {6},
  pages = {6142--6164},
  doi = {10.1137/26M1847996},
  url = {https://doi.org/10.1137/26M1847996},
  eprint = {2509.09088},
  archivePrefix = {arXiv},
}

@misc{aladrah2026geometric,
  author = {Nicola Aladrah and Emanuele Ballarin and Matteo Biagetti and Alessio Ansuini and Alberto d'Onofrio and Fabio Anselmi},
  title = {{Understanding and inverse design of implicit bias in stochastic learning: a geometric perspective}},
  year = {2026},
  url = {https://arxiv.org/abs/2601.06597v2},
  note = {Version 2, 4 April 2026},
  eprint = {2601.06597v2},
  archivePrefix = {arXiv},
  howpublished = {arXiv preprint},
}

@inproceedings{ziyin2024noiseequilibrium,
  author = {Liu Ziyin and Mingze Wang and Hongchao Li and Lei Wu},
  title = {{Parameter Symmetry and Noise Equilibrium of Stochastic Gradient Descent}},
  year = {2024},
  url = {https://arxiv.org/abs/2402.07193},
  booktitle = {Advances in Neural Information Processing Systems},
  volume = {37},
  pages = {93874--93906},
  eprint = {2402.07193},
  archivePrefix = {arXiv},
}

@inproceedings{ziyin2025thermodynamics,
  author = {Liu Ziyin and Yizhou Xu and Isaac Chuang},
  title = {{Neural Thermodynamics: Entropic Forces in Deep and Universal Representation Learning}},
  year = {2025},
  url = {https://arxiv.org/abs/2505.12387},
  booktitle = {Advances in Neural Information Processing Systems},
  volume = {38},
  note = {Published at NeurIPS 2025; arXiv version 4 revised 2 February 2026},
  eprint = {2505.12387},
  archivePrefix = {arXiv},
}

@inproceedings{chen2023stochasticcollapse,
  author = {Feng Chen and Daniel Kunin and Atsushi Yamamura and Surya Ganguli},
  title = {{Stochastic Collapse: How Gradient Noise Attracts SGD Dynamics Towards Simpler Subnetworks}},
  year = {2023},
  url = {https://arxiv.org/abs/2306.04251},
  booktitle = {Advances in Neural Information Processing Systems},
  volume = {36},
  eprint = {2306.04251},
  archivePrefix = {arXiv},
}

@article{lu2020dying,
  author = {Lu Lu and Yeonjong Shin and Yanhui Su and George Em Karniadakis},
  title = {{Dying ReLU and Initialization: Theory and Numerical Examples}},
  year = {2020},
  url = {https://arxiv.org/abs/1903.06733},
  journal = {Communications in Computational Physics},
  volume = {28},
  number = {5},
  pages = {1671--1706},
  doi = {10.4208/cicp.OA-2020-0165},
  eprint = {1903.06733},
  archivePrefix = {arXiv},
}

@article{watanabe2013wbic,
  author = {Sumio Watanabe},
  title = {{A Widely Applicable Bayesian Information Criterion}},
  year = {2013},
  url = {https://jmlr.org/papers/v14/watanabe13a.html},
  journal = {Journal of Machine Learning Research},
  volume = {14},
  number = {27},
  pages = {867--897},
}

@book{abramowitz1964handbook,
  url = {https://personal.math.ubc.ca/~cbm/aands/},
  editor    = {Milton Abramowitz and Irene A. Stegun},
  title     = {Handbook of Mathematical Functions with Formulas, Graphs, and Mathematical Tables},
  series    = {Applied Mathematics Series},
  volume    = {55},
  publisher = {National Bureau of Standards},
  address   = {Washington, DC},
  year      = {1964}
}

@inproceedings{gunasekar2018implicit,
  url = {https://proceedings.neurips.cc/paper_files/paper/2018/hash/0e98aeeb54acf612b9eb4e48a269814c-Abstract.html},
  title     = {Implicit Bias of Gradient Descent on Linear Convolutional Networks},
  author    = {Gunasekar, Suriya and Lee, Jason D. and Soudry, Daniel and Srebro, Nathan},
  booktitle = {Advances in Neural Information Processing Systems},
  volume    = {31},
  year      = {2018}
}

@inproceedings{du2018algorithmic,
  url = {https://proceedings.neurips.cc/paper/2018/hash/fe131d7f5a6b38b23cc967316c13dae2-Abstract.html},
  title     = {Algorithmic Regularization in Learning Deep Homogeneous Models:
               Layers are Automatically Balanced},
  author    = {Du, Simon S. and Hu, Wei and Lee, Jason D.},
  booktitle = {Advances in Neural Information Processing Systems},
  volume    = {31},
  year      = {2018}
}

@article{luther2020reexamining,
  url = {https://doi.org/10.1103/PhysRevResearch.2.033135},
  title   = {Reexamining the Principle of Mean-Variance Preservation for Neural Network Initialization},
  author  = {Luther, Kyle and Seung, H. Sebastian},
  journal = {Physical Review Research},
  volume  = {2},
  number  = {3},
  pages   = {033135},
  year    = {2020},
  doi     = {10.1103/PhysRevResearch.2.033135}
}

@article{BrascampLieb1976,
  url = {https://doi.org/10.1016/0022-1236(76)90004-5},
  author  = {Brascamp, Herm Jan and Lieb, Elliott H.},
  title   = {On extensions of the {Brunn--Minkowski} and {Pr{\'e}kopa--Leindler} theorems, including inequalities for log concave functions, and with an application to the diffusion equation},
  journal = {Journal of Functional Analysis},
  volume  = {22},
  number  = {4},
  pages   = {366--389},
  year    = {1976},
  doi     = {10.1016/0022-1236(76)90004-5}
}

@article{KolesnikovMilman2017,
  url = {https://doi.org/10.1007/s12220-016-9736-5},
  author  = {Kolesnikov, Alexander V. and Milman, Emanuel},
  title   = {{Brascamp--Lieb}-Type Inequalities on Weighted {Riemannian} Manifolds with Boundary},
  journal = {Journal of Geometric Analysis},
  volume  = {27},
  number  = {2},
  pages   = {1680--1702},
  year    = {2017},
  doi     = {10.1007/s12220-016-9736-5}
}

@inproceedings{zhu2019anisotropic,
  title     = {The Anisotropic Noise in Stochastic Gradient Descent:
               Its Behavior of Escaping from Sharp Minima and Regularization Effects},
  author    = {Zhu, Zhanxing and Wu, Jingfeng and Yu, Bing
               and Wu, Lei and Ma, Jinwen},
  booktitle = {Proceedings of the 36th International Conference on Machine Learning},
  series    = {Proceedings of Machine Learning Research},
  volume    = {97},
  pages     = {7654--7663},
  year      = {2019},
  publisher = {PMLR},
  url       = {https://proceedings.mlr.press/v97/zhu19e.html}
}

@inproceedings{haochen2021shape,
  title     = {Shape Matters: Understanding the Implicit Bias of the Noise Covariance},
  author    = {HaoChen, Jeff Z. and Wei, Colin and Lee, Jason and Ma, Tengyu},
  booktitle = {Proceedings of Thirty Fourth Conference on Learning Theory},
  series    = {Proceedings of Machine Learning Research},
  volume    = {134},
  pages     = {2315--2357},
  year      = {2021},
  publisher = {PMLR},
  url       = {https://proceedings.mlr.press/v134/haochen21a.html}
}

@article{aoyagi2005reducedrank,
  author  = {Aoyagi, Miki and Watanabe, Sumio},
  title   = {Stochastic complexities of reduced rank regression in {Bayesian} estimation},
  journal = {Neural Networks},
  year    = {2005},
  volume  = {18},
  number  = {7},
  pages   = {924--933},
  doi     = {10.1016/j.neunet.2005.03.014},
  url     = {https://doi.org/10.1016/j.neunet.2005.03.014}
}
\bibliographystyle{iclr2027_conference}

\appendix
\crefalias{section}{appendix}
\crefalias{subsection}{appendix}
\startcontents[appendices]

\printcontents[appendices]{l}{1}{\section*{Appendices}\setcounter{tocdepth}{2}}
\label{sec:appendix_toc}

\section{AI use disclosure}
Proof techniques were developed by the authors with AI assistance. AI suggestions were carefully checked and compared to existing literature, and the text and proofs themselves were written by the authors. AI was used to implement Langevin simulations, plot the resulting figures and draft descriptions of the experiments. We take responsibility for the final content of this work.

\section{Additional figures}

\subsection{Diagonal networks}
\begin{figure}[h]
    \centering
    \includegraphics[width=0.8\textwidth]{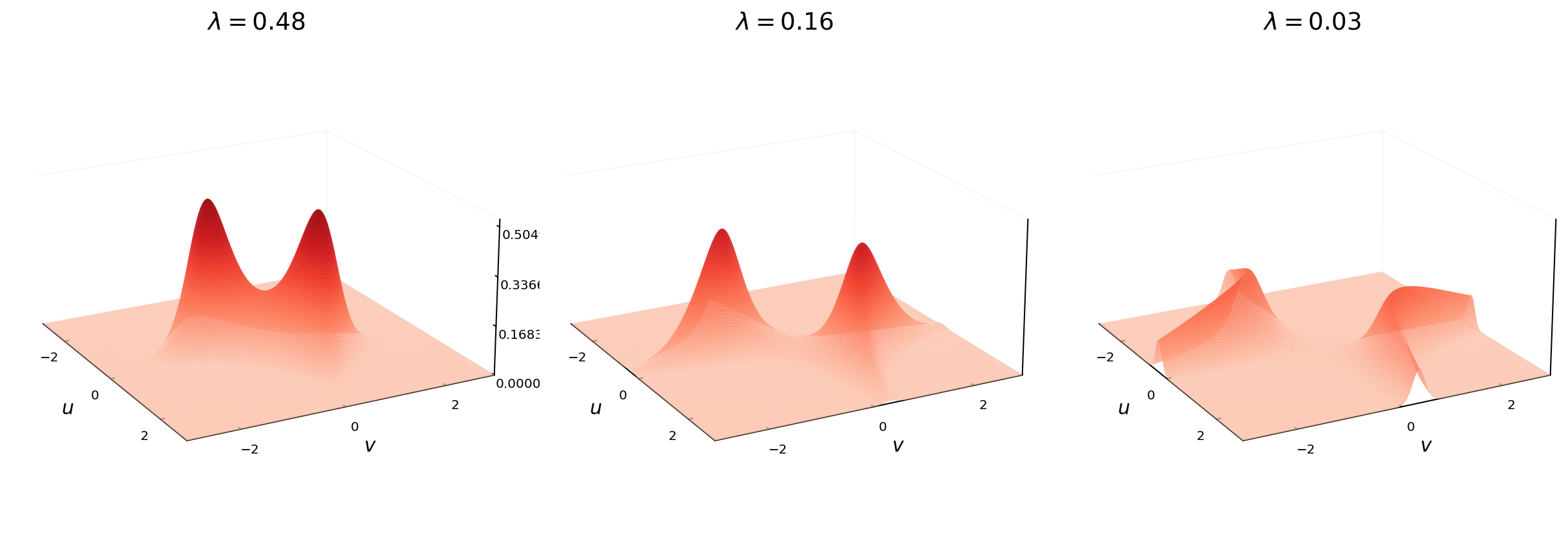}
    \caption{Stationary distributions for $w=uv$ in a two-layer diagonal network $f(x)=uvx$ trained by Langevin dynamics on the loss $L(u,v)=(uv-1)^2/2$ (temperature $T=0.15$). As $\lambda$ decreases, the density spreads across the loss valley, with mass escaping to infinity.}
    \label{fig:diag-uv-dist}
\end{figure}

\begin{figure}[h]
    \centering
    \includegraphics[width=0.7\textwidth]{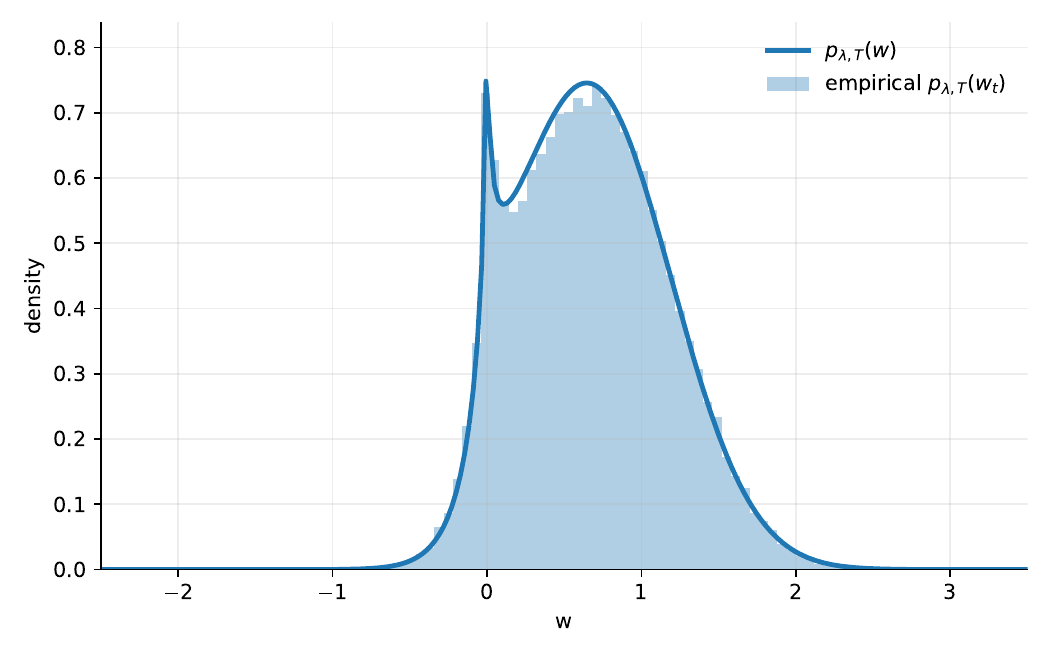}
    \caption{Analytical and empirical density of $w$ for one-layer diagonal network $f(x)=uvx$ with $L(u,v)=(uv-1)^2/2$ trained by Langevin dynamics, with $\lambda=0.2$ and $T=0.25$. Samples are collected after burn-in and thinned.}
    \label{fig:diag-exact}
\end{figure}

\begin{figure}[h]
    \centering
    \includegraphics[width=\textwidth]{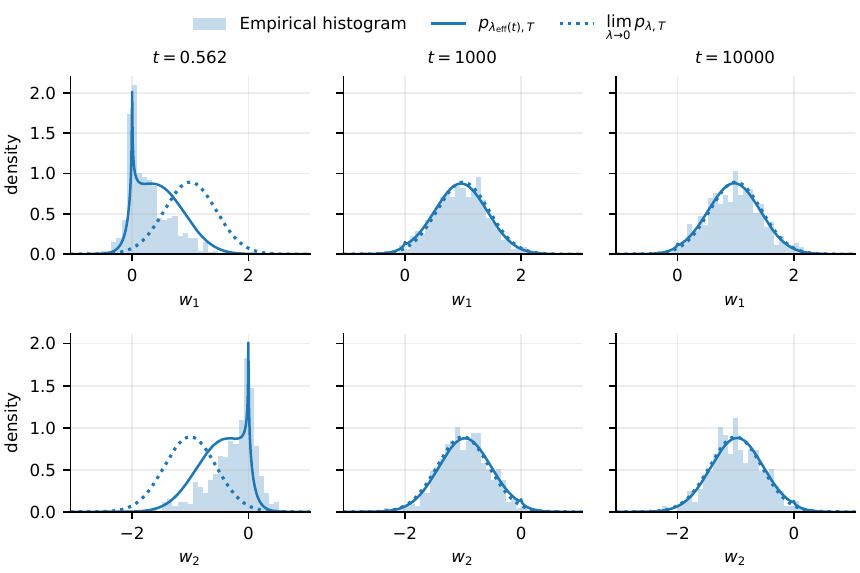}
    \caption{Densities for unregularized Langevin in the setting of \cref{fig:diag_alpha_eff,app:diag-experiment}, showing all results to time $t=10^4$.}
    \label{fig:diag-hist-long}
\end{figure}

\begin{figure}[h]
    \centering
    \includegraphics[width=0.8\textwidth]{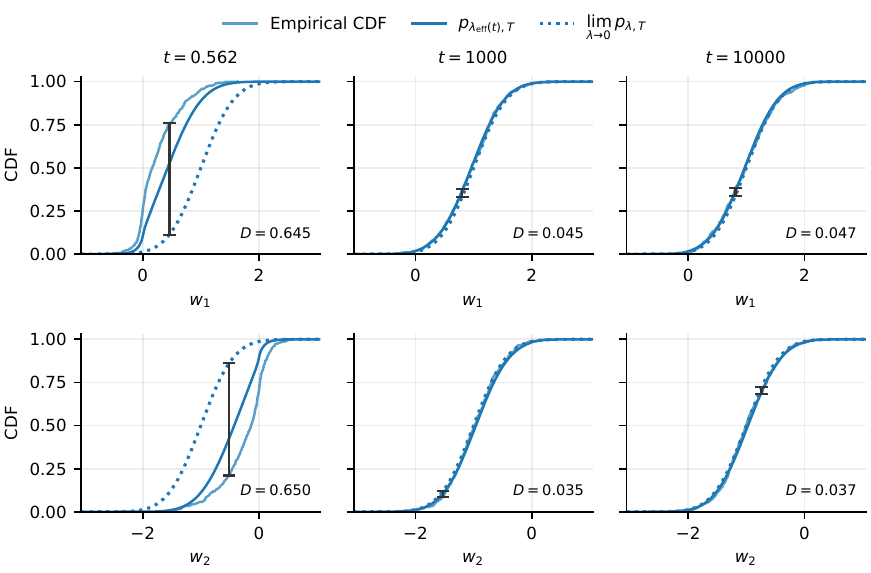}
    \caption{CDFs for unregularized Langevin in the setting of \cref{fig:diag_alpha_eff,app:diag-experiment}, up to time $t=10^4$. The vertical distance $D$ marks the largest disagreement between the empirical CDF and that of the predicted limit, that is, the Kolmogorov-Smirnov distance of the distributions.}
    \label{fig:diag-cdf}
\end{figure}

\begin{figure}[h]
    \centering
    \includegraphics[width=0.8\textwidth]{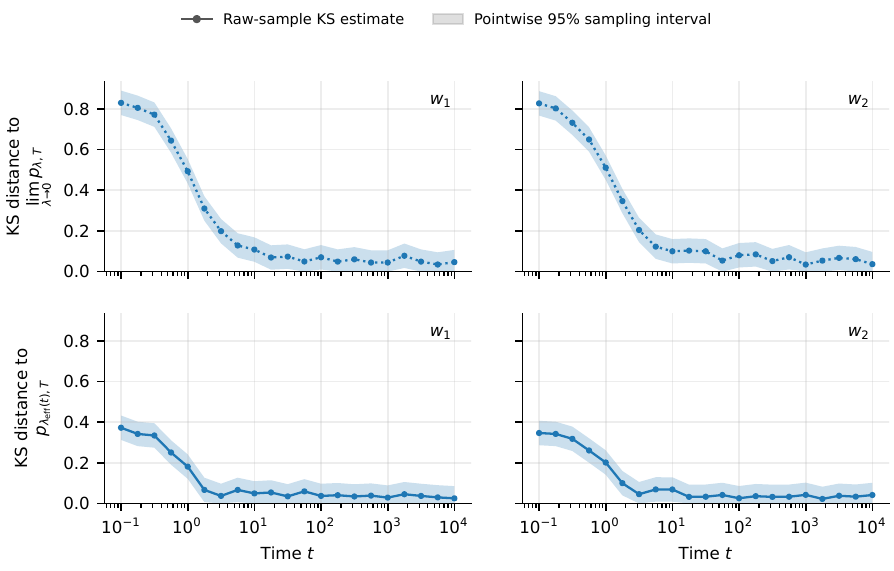}
    \caption{For the empirical CDF of unregularized Langevin in the setting of \cref{fig:diag_alpha_eff,app:diag-experiment}, we show its Kolmogorov-Smirnov distance to the predicted limit (top) and to the Gibbs distribution using $\lambda_\mathrm{eff}(t)$ (bottom). The latter shows a more rapid agreement as $t$ grows. There is little change in the distance to the predicted limit between $t=1000$ and $t=10^4$, suggesting the convergence from \cref{thm:diagonal} has been reached, up to numerical error.}
    \label{fig:diag-ks}
\end{figure}

\FloatBarrier

\subsection{Linear networks}

% \begin{figure}[h]
%     \centering
%     \includegraphics[width=\textwidth]{figures/finite_lambda_matrix.pdf}
%     \caption{Analytical and empirical density of the singular values of $W$ for a one-layer square linear network $f(x)=UV^\top x$ with $\ell(W)=\|W-W_*\|_F^2/2$ for $W=UV^\top$ and $W_*=\text{diag}(1.25, 0.55)$, trained by Langevin dynamics, with $\lambda=0.2$ and $T=0.25$. Samples are collected after burn-in and thinned.}
%     \label{fig:linear-exact}
% \end{figure}

\begin{figure}[h]
    \centering
    \includegraphics[width=\textwidth]{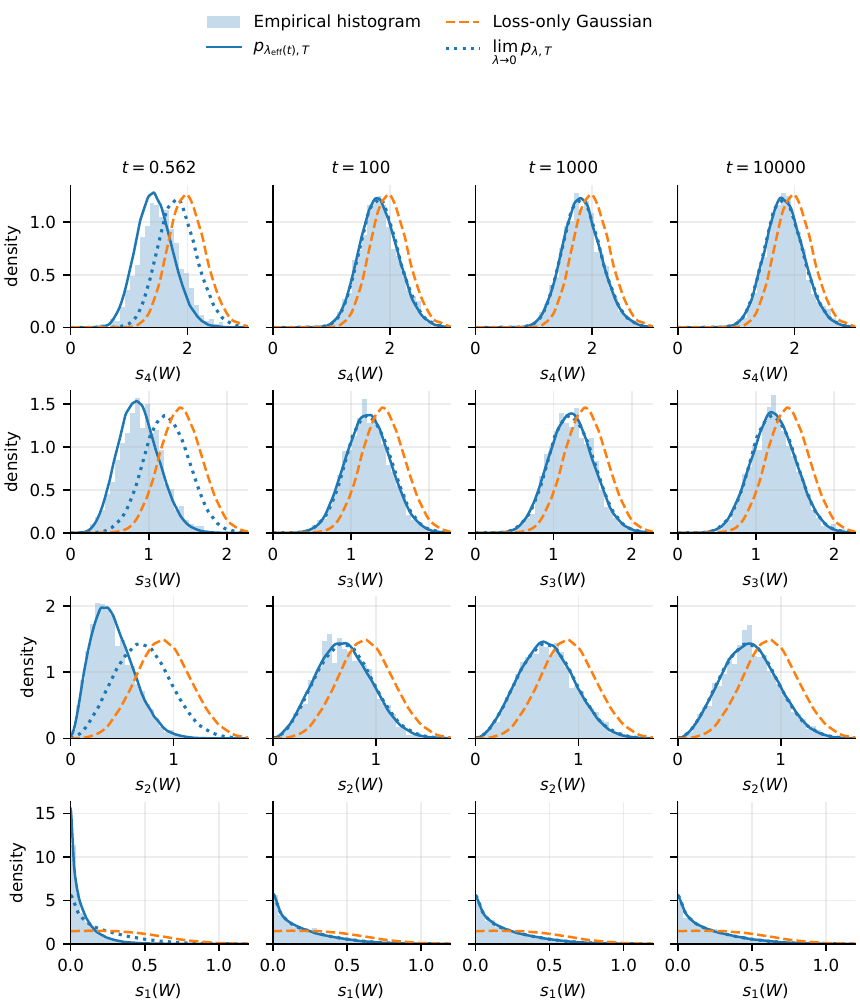}
    \caption{Marginal Langevin densities for all singular values in the setting of \cref{fig:linear_alpha_eff,app:linear-experiment}, including all time up to $t=10^4$. The predicted $\lambda\to0$ limit and empirical distribution appear to agree at large times, as do the empirical densities at time $t$ and Gibbs densities using $\lambda_\mathrm{eff}(t)$. There is a visible shift towards zero in the smaller singular values.}
    \label{fig:linear-histogram-long}
\end{figure}

\begin{figure}[h]
    \centering
    \includegraphics[width=\textwidth]{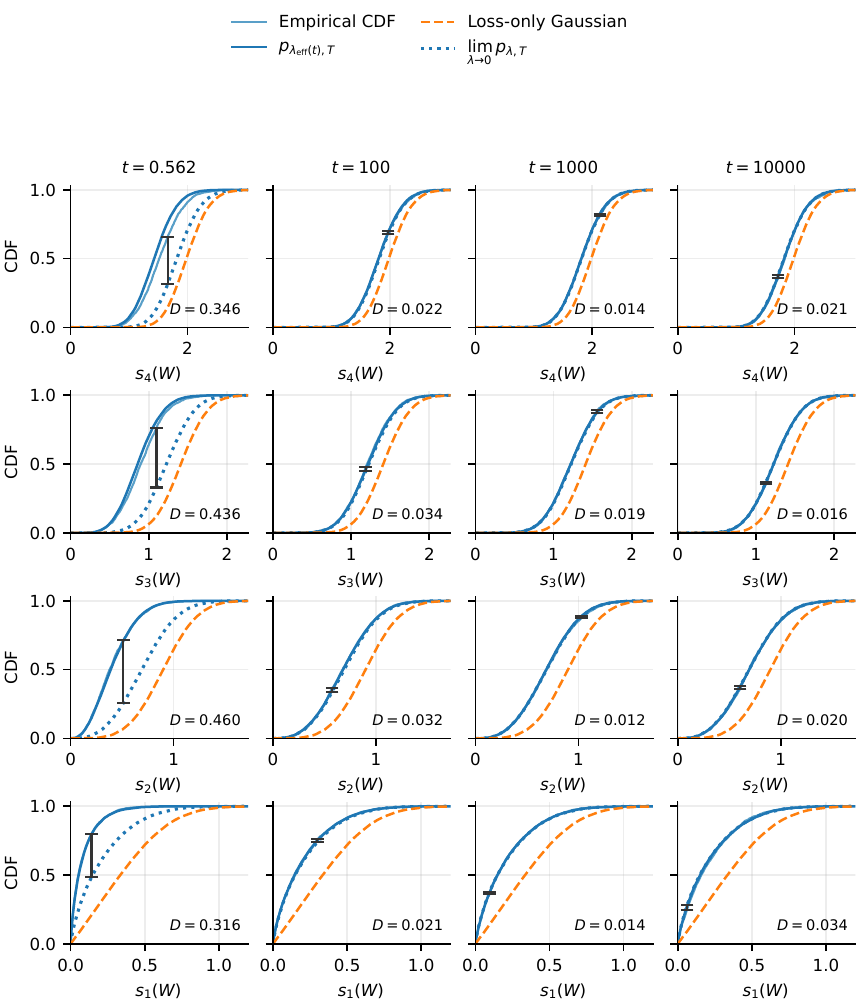}
    \caption{Marginal Langevin CDFs for all singular values in the setting of \cref{fig:linear_alpha_eff,app:linear-experiment}, including all time up to $t=10^4$. The vertical distance $D$ marks the Kolmogorov-Smirnov distance between the empirical distribution and predicted limiting distributions.}
    \label{fig:linear-cdf}
\end{figure}

\begin{figure}[h]
    \centering
    \includegraphics[width=\textwidth]{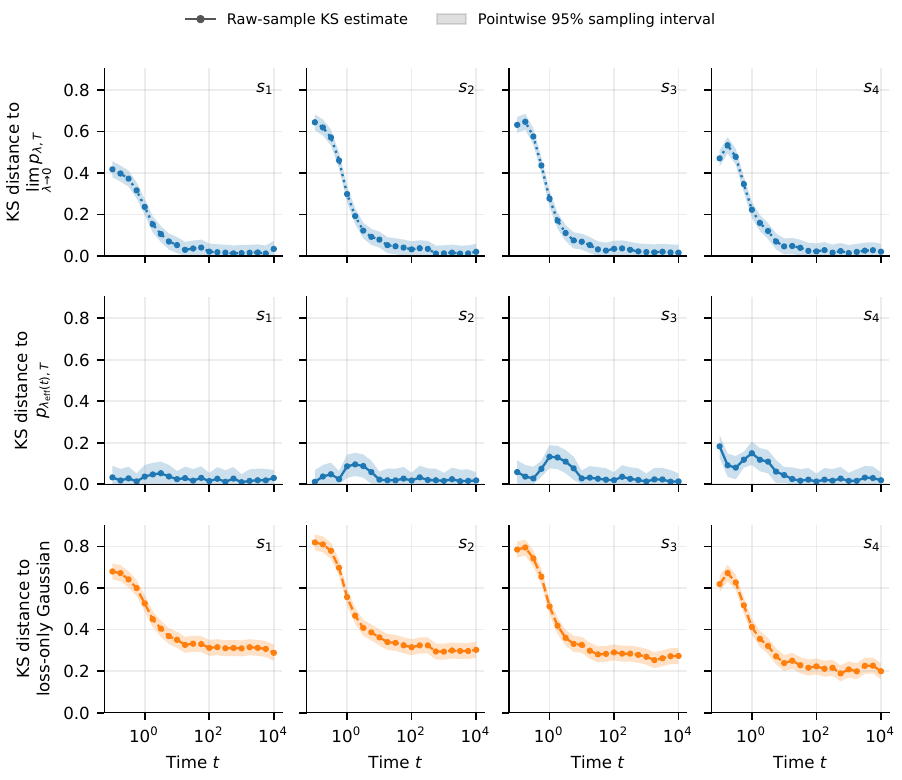}
    \caption{For the empirical CDF of unregularized Langevin in the setting of \cref{fig:linear_alpha_eff,app:linear-experiment}, we show its Kolmogorov-Smirnov distance to the predicted limit (top), to the Gibbs distribution using $\lambda_\mathrm{eff}(t)$ (middle), and to the na\"ive distribution proportional to $\exp(-\ell(W)/T)$ (bottom). The former two show agreement as $t$ grows, with the $\lambda_\mathrm{eff}(t)$ prediction being almost immediately accurate, while the na\"ive prediction does not match the data.  There is little change in the distance to the predicted limit between $t=1000$ and $t=10^4$, suggesting convergence.}
    \label{fig:linear-ks}
\end{figure}

\FloatBarrier

\subsection{ReLU networks}

\begin{figure}[h]
    \centering
    \includegraphics[width=\textwidth]{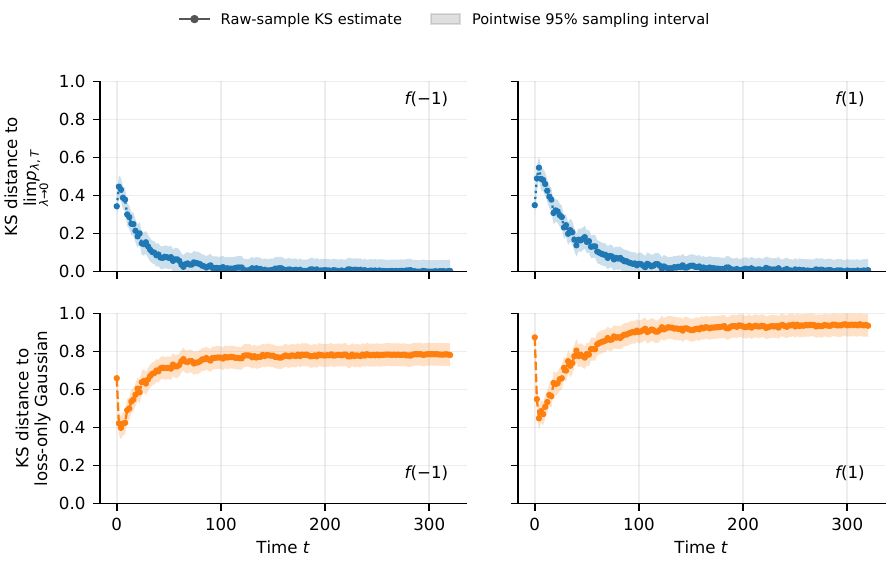}
    \caption{For the empirical CDF of unregularized Langevin in the setting of \cref{fig:relu_heat_death,app:heat-death-experiment}, we show its Kolmogorov-Smirnov distance of the distribution of outputs to the predicted zero limit (top) and to the na\"ive distribution proportional to $\exp(-\ell(f)/T)$ (bottom). The former shows agreement as $t$ grows, while the na\"ive prediction does not.  There is little change in the distance to the predicted limit between $t=200$ and $t=320$, suggesting convergence.}
    \label{fig:relu_ks}
\end{figure}

\FloatBarrier

\section{Proofs}
\subsection{Diagonal network integral limit}
\label{app:diag}

We examine the integral
\begin{align*}
Z_\alpha(w) = \frac{\log(1/\alpha)}{2} \int_{-\infty}^\infty e^{-\frac{|w|}{2}\left(\alpha^{1-s}+\alpha^{1+s}\right)}\,ds.
\end{align*}
As explained in the main text, the integrand converges to the indicator $\mathbf{1}_{(-1,1)}(s)$ as $\alpha\to0$. We apply the dominated convergence theorem to show the integral converges. Namely, for $|s|\le1$ the integrand is bounded by one, while for $|s|>1$, once $\alpha < e^{-1}$ the integrand is bounded by $\exp(-|w|e^{|s|-1}/2)$. The argument above generalizes readily to deep chains, and by the decoupling of coordinates, to general deep diagonal networks.

To formally obtain convergence of the distributions $p_{\lambda,T}(w)$ as $\lambda\to0$, one needs an upper bound integrable against $e^{-\ell(w)/T}\,dw$. Such a bound holds, for example, for convex coercive losses such as the quadratic losses used in our examples and experiments. Namely, for a depth-$D$ chain the upper bound
\begin{align*}
\frac{Z_\alpha(w)}{\log(1/\alpha)^{D-1}} \lesssim(1+ |\log|w||^{D-1})
\end{align*}
is valid once $\alpha\le e^{-1}$.
Keeping track of one more logarithmic order (for example via Meijer-G functions, as in \citet{zavatoneveth2021exact}) gives the expansion
\begin{align*}
 Z_\alpha(w)
 =C_D \log(1/\alpha)^{D-1}
 +O\left(\log(1/\alpha)^{D-2}
    (1+|\log|w||^{D-1})\right)
\end{align*}
for a constant $C_D>0$, which we use in the proof of the interchange of limits for diagonal networks.

\subsection{
Diagonal network limit interchange}
\label{app:diag-lim-int}
The effective potential from the main text generalizes to $\lambda\ge0$ by considering the diffusive scale
\begin{align*}
\frac{d}{dt} r_\lambda(t)^2 = 2T-2\lambda r_\lambda(t)^2,
\qquad
r_\lambda(0)^2=s_0^2>0.
\end{align*}
The choice of $s_0^2$ is free, though it is natural to choose the initial scale of $\theta$; in our experiments, initialization is a centered Gaussian, with $s_0^2$ its variance. Consider the rescaling of parameters $X_\tau = \theta_t/r_\lambda(t)$, and time $d\tau = dt/r_\lambda(t)^2$. By It\^o's formula, $X_\tau$ follows Langevin dynamics
\begin{align*}
dX_\tau = -\nabla V_\tau(X_\tau)\,dt + \sqrt{2T}\,dB_\tau,
\qquad V_\tau(X)=L(r_\lambda(t(\tau))X) + \frac{T\|X\|^2}{2},
\end{align*}
with instantaneous potential $V_\tau$. If we consider freezing this potential at time $\tau$, transforming back to $\theta$ gives the potential $L(\theta)+\lambda_\mathrm{eff}(t(\tau),\lambda)R(\theta)$ with effective regularization
\begin{align*}
\lambda_\mathrm{eff}(t,\lambda)=\frac{T}{r_\lambda(t)^2},
\qquad
r_\lambda(t)^2 = 
\begin{cases}
    s_0^2 e^{-2\lambda t} + \frac{T}{\lambda}(1-e^{-2\lambda t}) & \lambda>0\\
    s_0^2 + 2Tt & \lambda=0.
\end{cases}
\end{align*}
We restate the theorem here.
\diagthm*
For $\lambda=0$, the effective regularization vanishes as $t\to\infty$, and thus $\Pr_{\lambda_\mathrm{eff}(t,0),T}(dw)$ approaches $\lim_{\lambda\to0} \Pr_{\lambda,T}(dw)$ in total variation, and
$\lim_{t\to\infty} \Pr_{t,0,T}(dw)= \lim_{\lambda\to0}\Pr_{\lambda,T}(dw)$. Under the uniform bound conditions, $\|\Pr_{t,0,T}(dw)-\Pr_{\lambda_\mathrm{eff}(t,0),t,T}\|_\mathrm{TV}\lesssim\log(T/\lambda_\mathrm{eff}(t,0))^{-1/2}$, while \cref{app:diag} shows $\|\Pr_{\lambda_\mathrm{eff}(t,0),t,T}-\lim_{\lambda\to0}P_{\lambda,T}\|_{\mathrm{TV}}=O(\log(T/\lambda_\mathrm{eff}(t,0))^{-1})$. A triangle inequality shows $\|\Pr_{t,0,T}(dw)-\lim_{\lambda\to0}P_{\lambda,T}\|_\mathrm{TV}\lesssim\log(T/\lambda_\mathrm{eff}(t,0))^{-1/2}$

The proof rests on two claims. The first concerns equilibration under the effective potential $V_\tau$. Freezing $\tau$ fixes a value of $r$, and we denote the resulting Gibbs distribution of $X_\tau$ under $V_\tau$ by
\begin{align*}
\mu_r(dx) \propto \exp\left(-\frac{L(rx)}{T} - \frac{\|x\|^2}{2}\right)\,dx.
\end{align*}
Pushing $\mu_r$ forward by $x\mapsto rx$ recovers $P_{\lambda_\mathrm{eff}(\tau),T}$ with $\lambda_\mathrm{eff}(\tau)=\lambda_\mathrm{eff}(t(\tau),\lambda)$.

\begin{lemma}[Uniform Poincar\'e inequality]
\label{lem:poincare}
There exists a constant $K<\infty$ independent of $r$ and depths $D_i$ such that for any  $r>0$ and test functions $g \in H^1(\mu_r)$,
\begin{align*}
\Var_{\mu_r}(g) \le K \int \|\nabla g\|^2\,d\mu_r.
\end{align*}
\end{lemma}

\begin{lemma}[Dissipation]
\label{lem:dissip}
Let $h_\tau$ be the probability density of $X_\tau$ with respect to $\mu_r$, and define its $\chi^2$ divergence from uniformity $D_\tau = \int (h_\tau-1)^2\,d\mu_r$. This measures how far $X_\tau$ is from equilibrium under $\mu_r$. For sufficiently small $\lambda$, for almost every $\tau$
\begin{align*}
D_\tau' \le -T\left(\frac{\lambda}{\lambda_\mathrm{eff}(\tau)}+1\right) \int \|\nabla h_\tau\|^2\,d\mu_r + T\left(1-\frac{\lambda}{\lambda_\mathrm{eff}(\tau)}\right)(1+D_\tau)\E_{\mu_r}\|X\|^2.
\end{align*}
\end{lemma}

We first prove the theorem assuming these two lemmas.
\begin{proof}[Proof of \cref{thm:diagonal}]
We first prove the quantitative bound with $p_{0,\lambda,T}\in L^2$ and $\E_{P_{0,\lambda,T}}[\|\theta\|+\sqrt{L(\theta)-\min\ell}]<\infty$, the result for arbitrary initializations following by approximation.  We claim it suffices to prove the bound under a particular initial distribution with finite $\chi^2$ divergence from $P_{\lambda_\mathrm{eff}(\tau),T}$. In particular, given $t$, let $b=\log r_\lambda(t)^2$, and let
\begin{align*}
E_0(\theta)=\frac{L(\theta)-\min\ell}{T}+\frac{\|\theta\|^2}{2s_0^2}.
\end{align*}
For a constant $a>0$ to be specified, define the truncated initial distribution
\begin{align*}
Q_b(\cdot)=P_{0,\lambda,T}(~\cdot \mid E_0(\theta)\le ab).
\end{align*}
The total variation distance between $P_{0,\lambda,T}$ and $Q_b$ is then bounded by
\begin{align*}
P_{0,\lambda,T}(E_0(\theta)> ab) \le \frac{\E_{0,\lambda,T}\sqrt{E_0(\theta)}}{\sqrt{ab}}=O\left(b^{-1/2}\right),
\end{align*}
applying Markov's inequality and the assumption on initialization. Since Langevin dynamics contract total variation, by the triangle inequality it suffices to prove the result with $P_{0,\lambda,T}=Q_b$. Taking $b$ large enough, this initialization enjoys controlled initial $\chi^2$ divergence,
\begin{align*}
1+D_{\chi^2}(Q_b\|P_{\lambda_\mathrm{eff}(0),T}) \le C\int_{E_0\le ab} e^{E_0(\theta)} p_{0,\lambda,T}(\theta)^2\,d\theta\le Ce^{ab}\|p_{0,\lambda,T}\|^2_{L^2}.
\end{align*}

Assume such an initial distribution. In the notation of \cref{lem:dissip}, $\Var_{\mu_r}(h_\tau)=D_\tau$, from which \cref{lem:poincare} implies $\frac{D_\tau}{K}\le\int\|\nabla h_\tau\|^2 \,d\mu_r$.
Plugging this into the bound in \cref{lem:dissip} gives
\begin{align*}
D_\tau'\le -T\left(\frac{1+\lambda / \lambda_\mathrm{eff}(\tau)}{K} - \left(1-\frac{\lambda}{\lambda_\mathrm{eff}(\tau)}\right)\E_{\mu_r}\|X\|^2\right)D_\tau
+ T \left(1-\frac{\lambda}{\lambda_\mathrm{eff}(\tau)}\right)\E_{\mu_r}\|X\|^2
\end{align*}
Writing the partition function of $P_{\lambda_\mathrm{eff}(\tau),T}$ as $\mathcal{Z}(r)$, the expansion from \cref{app:diag} gives
\begin{align*}
\mathcal{Z}(r) = \int e^{-L(\theta)/T-\|\theta\|^2/2r^2}\,d\theta 
= C(\log r^2)^{-n+\sum_iD_i} + O\left((\log r^2)^{-n-1+\sum_i D_i}\right).
\end{align*}
We obtain a bound on $\E_{\mu_r}\|X\|^2$ by writing
\begin{align*}
\frac{\mathcal{Z}(\sqrt2r)}{\mathcal{Z}(r)} = \frac{1}{{\mathcal{Z}(r)}}\int e^{\|\theta||^2/4r^2}e^{-L(\theta)/T-\|\theta\|^2/2r^2}\,d\theta=\E_{\mu_r}e^{\|X\|^2/4} \ge 1+\frac{\E_{\mu_r}\|X\|^2}{4}.
\end{align*}
Dividing the asymptotic expansions, and enlarging $K$ if necessary, $\E_{\mu_r}\|X\|^2\le K/\log(r^2)$ for sufficiently large $r$. It follows that
\begin{align*}
D_\tau' \le -\frac{T}{2K}D_\tau + \frac{TK}{\log r^2}\left(1-\frac{\lambda}{\lambda_\mathrm{eff}(\tau)}\right).
\end{align*}
The first term shows $X_\tau$ equilibrating under $V_\tau$, while the second measures the change in the potential, with the term in parenthesis being less than one for small enough $\lambda$. Writing $B_\tau = \log r(t(\tau))^2$, we have from the definition of $r(t)$ that
\begin{align*}
B_\tau'=2T\left(1-\frac{\lambda}{\lambda_\mathrm{eff}(\tau)}\right) \le 2T,
\end{align*}
the upper bound holding for $\lambda s_0^2\le T$, when the factor above is in $[0,1]$. For $B_\tau \ge 8K^2$, we obtain
\begin{align*}
(B_\tau D_\tau)'\le-\frac{T}{4K}B_\tau D_\tau + TK.
\end{align*}
Letting $B_*$ be large enough for this to hold, and $B_{\tau_*}=B_*$, integrating up to a later time $\tau$ gives, by the boundedness of $B_\tau'$,
\begin{align*}
B_\tau D_\tau \le 4K^2 + B_* D_{\tau_*} e^{-T(\tau-\tau_*)/4K}.
\end{align*}
To bound the contribution before $\tau_*$, note from \cref{lem:dissip} that
\begin{align*}
D_\tau' \le T\left(1-\frac{\lambda}{\lambda_\mathrm{eff}(\tau)}\right)\E_{\mu_r}\|X\|^2(1+D_\tau).
\end{align*}
The factor in front of $(1+D_\tau)$ equals $\frac{d}{d\tau} \log \mathcal{Z}(r)$. Integrating up to $\tau_*$ gives
\begin{align*}
1+D_{\tau_*} \le \frac{\mathcal{Z}(e^{B_*/2})}{\mathcal{Z}(s_0)}(1+D_0).
\end{align*}
Furthermore, $\tau-\tau_*\ge (B_\tau-B_*)/(2T)$ since $B_\tau'\le 2T$, so we get
\begin{align*}
B_\tau D_\tau \le 4K^2 + C(1+D_0)e^{-B_\tau/8K}.
\end{align*}
We now apply this bound to our initialization $Q_b$ with $b=B_{\tau(t)}$ for our fixed $t$. Recall that
\begin{align*}
1+D_0 = 1+D_{\chi^2}(Q_b\|P_{\lambda_\mathrm{eff}(0),T}) \le Ce^{ab}\|p_{0,\lambda,T}\|^2_{L^2}.
\end{align*}
Choosing $a=1/(16K)$, and absorbing the squared norm into $C$, the previous bound gives
\begin{align*}
b D_\tau \le  4K^2 + Ce^{ab-b/8K} =  4K^2 + Ce^{-b/16K} \le C.
\end{align*}
Finally, this means
\begin{align*}
\|P_{t,\lambda,T}-P_{\lambda_\mathrm{eff}(t,\lambda),T}\|_\mathrm{TV} \le \frac{1}{2} \sqrt{D_\tau} = O\left(b^{-1/2}\right).
\end{align*}
Since $b=\log r_\lambda(t)^2=\log(T/\lambda_\mathrm{eff}(t,\lambda))$, the quantitative bound is proved.
For arbitrary fixed initializations, the non-quantitative limit is obtained by approximating the distribution after some fixed time by a compactly supported distribution with smooth density, and applying the result above together with contraction and continuity and thus uniformity in $\lambda$.
\end{proof}

It remains to prove the two lemmas above in turn.

\begin{proof}[Proof of \cref{lem:poincare}]
Let $G$ be the standard Gaussian measure on parameter space, and for $b_i > 0$ let
\begin{align*}
A_b = \left\{((x_{di})_{d=1}^{D_i})_{i=1}^n : \left|\prod_{d} x_{di}\right| \le b_i~\forall i\right\}.
\end{align*}
We show that $G(~\cdot\mid A_b)$ has Poincar\'e constant at most 1, before extending the result to $\mu_r$. Note that $A$ is invariant under sign flips $x_{di}\mapsto -x_{di}$. By symmetry of $G$, the test function $g$ decomposes into components of even and odd parity orthogonal under $G$ (with gradients orthogonal in Dirichlet energy). It thus suffices to find a uniform bound across the components. A component odd in $x_{di}$ has zero mean conditionally on the other variables, since $x_{di}$ is conditionally a symmetrically truncated Gaussian. Any odd component thus has Poincar\'e constant bounded by one. For even component, by symmetry, it suffices to consider the positive orthant (all $x_{di}>0$). In the coordinates $y_{di}=\log x_{di}$, the set $A_b=\{\sum_{di} y_{di}\le \log b_i\}$ is convex. Writing the density in the new coordinates as proportional to $\exp(-U(y))$, with $U(y)=\sum_j e^{2y_j}/2 - y_j$, the convex-boundary Brascamp-Lieb inequality \citep{BrascampLieb1976,KolesnikovMilman2017} gives a Poincar\'e constant bounded by $1/2$.

We extend the result to $\mu_r(dx) \propto \exp(-\ell(w)/T -\|x\|^2/2)\,dx$, where $w_i =r^{D_i}\prod_d x_{di}$. Taking the region $A_b$ with $b_i = r^{-D_i}M$, the Holley–Stroock bounded perturbation inequality gives
\begin{align*}
\Var_{\mu_r(~\cdot\mid A_b)}(g)\le e^{\omega_M(\ell)/T}\int\|\nabla g\|^2 \,d\mu_r(~\cdot\mid A_b),
\qquad
\omega_M(\ell)=\max_{[-M,M]^n}\ell-\min_{[-M,M]^n}\ell.
\end{align*}
It remains to extend this inequality to the entire space. To do so, let
\begin{align*}
S(x)=\sum_{i=1}^n \frac{1}{D_i}\sum_{d=1}^{D_i} x_{di}^2,
\qquad
W(x) = 1 + S.
\end{align*}
Let $\mathcal A$ be the generator of the process with equilibrium $\mu_r$. Direct calculation gives
\begin{align*}
\mathcal{A}S = 2Tn-2w\cdot\nabla\ell(w)-2TS.
\end{align*}
By convex coercivity, $w\cdot\nabla\ell(w)\ge c\|w\|-C$ for some $c>0$ and $C<\infty$. Therefore, for large enough $M$,
\begin{align*}
-\frac{\mathcal AW}{W} \ge -C_0 ~\text{ on $A_b$},
\qquad
-\frac{\mathcal AW}{W} \ge T ~\text{ on $A_b^c$},
\end{align*}
with the constants independent of $r$ and the depths $D_i$. By reversibility of $\mathcal A$, completing the square gives for any smooth compactly supported $u$
\begin{align}
\int -\frac{\mathcal AW}{W} u^2\,d\mu_r = T \int (2u \nabla u\cdot\nabla\log W - u^2\|\nabla\log W\|^2)\,d\mu_r
\le T\int\|\nabla u\|^2\,d\mu_r.
\end{align}
Taking $u=g-\E_{\mu_r(~\cdot\mid A_b)} g$, we get $\Var_{\mu_r}(g)\le\int u^2\,d\mu_r$. Therefore,
\begin{align*}
\Var_{\mu_r}(g)\le \int\|\nabla g\|^2\,d\mu_r +  \left(1+\frac{C_0}{T}\right)\int_{A_b} u^2\,d\mu_r
\le
\left(1+\left(1+\frac{C_0}{T}\right)e^{\omega_M(\ell)/T}\right)\int\|\nabla g\|^2\,d\mu_r,
\end{align*}
and spatial cutoffs extend the inequality to $g\in H^1$.
\end{proof}

We conclude with the proof of the second lemma, using notation from the proof of \cref{thm:diagonal}.
\begin{proof}[Proof of \cref{lem:dissip}]
To begin, we assume $h_\tau$ is nice enough to allow integration by parts and differentiation under the relevant integrals. Write the density of $X_\tau$ as $\rho_\tau=h_\tau\mu_{r_\lambda(t(\tau))}$, with a slight abuse of notation treating $\mu_r$ as a probability density. The Fokker-Planck equation for the rescaled dynamics with potential $V_\tau$ reads
\begin{align*}
    \partial_\tau\rho_\tau=T\nabla\cdot(\mu_r\nabla h_\tau).
\end{align*}
Recalling $D_\tau=\int (h_\tau-1)^2\,d\mu_r$, differentiating and integrating by parts,
\begin{align*}
    D_\tau'=-2T\int\|\nabla h_\tau\|^2\,d\mu_r-\int h_\tau^2\partial_\tau\log\mu_r\,d\mu_r.
\end{align*}
The first term corresponds to dissipation under $V_\tau$, and the second to the change of $V_\tau$.
Note that $(\log r_\lambda(\tau))'=T(1-\lambda/\lambda_\mathrm{eff}(\tau))$, so
\begin{align*}
    \partial_\tau\log\mathcal{Z}=T\left(1-\frac{\lambda}{\lambda_\mathrm{eff}(\tau)}\right)\E_{\mu_r}\|X\|^2
\end{align*}
and the log-density evolves as
\begin{align*}
\partial_\tau \log \mu_r(x) = \left(1-\frac{\lambda}{\lambda_\mathrm{eff}(\tau)}\right)\left(T(N-\E_{\mu_r}\|X\|^2)-x\cdot\nabla_x L(rx)\right).
\end{align*}
Substituting this into the equation for $D_\tau'$, and using
\begin{align*}
\int h_\tau^2x\cdot\nabla_xL(rx)\,d\mu_r=T\int\left(\left(N-\|x\|^2\right)h_\tau^2+2h_\tau x\cdot\nabla h_\tau\right)\,d\mu_r,
\end{align*}
completing the square gives
\begin{align*}
D_\tau'=&-T\left(\frac{\lambda}{\lambda_\mathrm{eff}(\tau)}+1\right)\int\|\nabla h_\tau\|^2\,d\mu_r
-T\left(1-\frac{\lambda}{\lambda_\mathrm{eff}(\tau)}\right)\int\|\nabla h_\tau-xh_\tau\|^2\,d\mu_r\\
&+T\left(1-\frac{\lambda}{\lambda_\mathrm{eff}(\tau)}\right)(1+D_\tau)\E_{\mu_r}\|X\|^2.
\end{align*}
Dropping the middle term, which is non-positive for small $\lambda$, gives the desired bound. To extend to arbitrary initialization with finite $D_0$, consider the modified process killed on reaching the ball of radius $C>0$. Its subdensity $h_{\tau,C}$ relative to $\mu_r$ and the corresponding $D_{\tau,C}=\int h_{\tau,C}^2\,\mu_r-1$ then obey the inequality proved above, with uniform bounds on $\|h_{\tau,C}\|_{L^2(\mu_r)}^2$ and $\int_0^{\tau_1}\|\nabla \,h_{\tau,C}\|_{L^2(\mu_r)}^2\,d\tau$. But one may show that $X_\tau$ is non-explosive, using the Lyapunov function $S$ from the proof of \cref{lem:poincare}. The distribution of killed process thus converges to that of $X_\tau$, and monotone convergence for $\|h_{\tau,C}\|_{L^2(\mu_r)}^2$ and lower semicontinuity for $\|\nabla h_{\tau,C}\|_{L^2(\mu_r)}^2$ give the integrated bound on any time interval. The almost-everywhere bound follows.
\end{proof}

\subsection{Two layer linear network entropy}
\label{app:linear-entropy}

Recall that
\begin{align*}
Z_\alpha(S) \propto \int_{X\succ 0} e^{-\frac{\alpha}{2}\Tr\left(S \left(X+  X^{-1} \right)\right)}|X|^{-(n+1)/2} \,dX.
\end{align*}
The scaling argument proceeds by compactification, or evaluation at a boundary, of the positive definite cone. We begin by identifying which boundary dominates, i.e.\ how many eigenvalues of $X$ go to zero in the dominant integrand. We then calculate the limiting fiber partition function at an arbitrary boundary, and confirm that the one we identify is indeed dominant.

Let $X$ have eigenvalues $\lambda_i(X)$. The two trace terms in $Z_\alpha(S)$ define an approximate cutoff region for these eigenvalues. Since $S$ is fixed and positive definite, $\Tr(SX) \sim \Tr(X)$, so the first trace term cuts off the integral when $\alpha \Tr(X)\gg1$. Similarly, $\Tr(SX^{-1})\sim \Tr(X^{-1})$ and so the second term cuts off the integral when $\alpha\Tr(X^{-1})\gg 1$. Therefore, the integral is non-negligible on the region
\begin{align*}
\alpha\lesssim \lambda_i(X) \lesssim \frac{1}{\alpha}.
\end{align*}
Intuitively, the dominant contribution comes from having some number $k$ of eigenvalues of order $\alpha$, and $r=n-k$ of order $\alpha^{-1}$, and is of order $\alpha^{-kr}$, which in turn is maximized when $k=\floor{n/2}$. The argument below formalizes this intuition.

It is convenient to change variables to $Y=\alpha S^{1/2} X S^{1/2}$, so that $Z_\alpha(S) \propto I_{\alpha^2}(S^2)$ where
\begin{align*}
I_\epsilon(D) = \int_{Y\succ 0} e^{-\frac{1}{2}\Tr Y} e^{-\frac{\epsilon}{2} \Tr\left(D Y^{-1}\right)}|Y|^{-(n+1)/2}\,dY.
\end{align*}
The relevant range is $\epsilon \lesssim \lambda_i(Y) \lesssim 1$, and we study the case where $k$ eigenvalues have scale $\epsilon$ and $r$ have scale $1$. Any positive definite $Y$ can be written
\begin{align*}
Y =
\begin{pmatrix}
    I_r     &   0   \\
    B^\top  &   I_k
\end{pmatrix}
\begin{pmatrix}
    A_1     &   0   \\
    0       &   \epsilon A_0
\end{pmatrix}
\begin{pmatrix}
    I_r     &   B   \\
    0       &   I_k
\end{pmatrix}
\end{align*}
for positive definite $A_1 \in \mathbf{S}^r_{++}$ and $A_0 \in \mathbf{S}^k_{++}$, and arbitrary $B \in \R^{r\times k}$. Here $A_1$ and $A_0$ correspond to the order-$1$ and order-$\epsilon$ directions, respectively, and $B$ describes the orientation of the vanishing directions. When $\epsilon =0$, the kernel of $Y$ is the image of
\begin{align*}
N_B =
\begin{pmatrix}
    -B \\
    I_k
\end{pmatrix}.
\end{align*}
The determinant is then $|Y|=\epsilon^k|A_1||A_0|$, and the change of measure is
\begin{align*}
dY = \epsilon^{k(k+1)/2} |A_1|^k \,dA_1 \,dA_0 \,dB.
\end{align*}
Therefore,
\begin{align*}
|Y|^{-(n+1)/2}\,dY =
\epsilon^{-kr/2}|A_1|^{k-(n+1)/2}|A_0|^{-(n+1)/2}\,dA_1 \,dA_0 \,dB,
\end{align*}
and we see indeed that for $\epsilon=\alpha^2$, the order of the integral is $\alpha^{-kr}$.

In order to evaluate the integral, note that
\begin{align*}
\Tr Y &= \Tr(A_1(I_r+BB^\top)) + \epsilon\Tr A_0,\\
\epsilon \Tr DY^{-1} &= \Tr(N_B^\top D N_B A_0^{-1}) + \epsilon \Tr(D_{11} A_1^{-1}),
\end{align*}
where $D_{11}$ is the upper-left $r$-by-$r$ block of $D$.
Therefore, $I_\epsilon(D)$ is the integral with respect to $dB$ of the product of two integrals:
\begin{align*}
\int_{A_1\succ 0} e^{-\frac{1}{2} \Tr(A_1(I_r+BB^\top))  - \frac{\epsilon}{2}\Tr(D_{11} A_1^{-1})}|A_1|^{k-(n+1)/2}\,dA_1
\end{align*}
and
\begin{align*}
\int_{A_0\succ0} e^{-\frac{1}{2} \Tr(N_B^\top D N_B A_0^{-1}) -\frac{\epsilon}{2} \Tr A_0}|A_0|^{-(n+1)/2}\,dA_0.
\end{align*}
Consider the integral over $A_1$, which using $n=k+r$ we write as
\begin{align*}
\int_{A_1\succ 0} e^{-\frac{1}{2} \Tr(A_1(I_r+BB^\top))  - \frac{\epsilon}{2}\Tr(D_{11} A_1^{-1})}|A_1|^{\frac{k}{2}-\frac{r+1}{2}}\,dA_1.
\end{align*}
The first exponential trace term keeps the integral finite as $A_1\to\infty$. The second exponential trace term gives the cutoff $\epsilon \lesssim A_1 \lesssim 1$, and so the integral behaves like
\begin{align*}
\int_\epsilon^1 a^{\frac{k}{2}-\frac{r+1}{2}} \,da.
\end{align*}
This is finite in the $\epsilon\to 0$ limit exactly when the exponent is greater than $-1$, that is, when $k\ge r$. When $k=r-1$ the integral is order $\log(1/\epsilon)$, as we show below. If $k<r-1$, then the integral diverges like a power of $\epsilon$, meaning our scaling is missing some directions which are in fact small. Similarly, the integral over $A_0$, changing variables to $H=A_0^{-1}$, is
\begin{align*}
\int_{H\succ 0}e^{-\frac{1}{2} \Tr(N_B^\top D N_B H) - \frac{\epsilon}{2}\Tr(H^{-1})}|H|^{\frac{r}{2}-\frac{k+1}{2}}\,dH.
\end{align*}
This is finite in the $\epsilon\to0$ limit when $r\ge k$, and order $\log(1/\epsilon)$ when $r=k-1$. We thus see that $k=\floor{n/2}$ and $r=\ceil{n/2}$ is the dominant scaling.

To evaluate the integrals, recall the normalization integral for an $m$-by-$m$ Wishart matrix $H$ with $n$ degrees of freedom and scale parameter $M \in \mathbf{S}^m_{++}$,
\begin{align*}
\int_{H\succ 0} e^{-\frac{1}{2}\Tr(M^{-1} H)} |H|^{\frac{n}{2}-\frac{m+1}{2}}\,dH = 2^{nm/2}\Gamma_m\left(\frac{n}{2}\right)|M|^{n/2},
\end{align*}
where $\Gamma_m$ is the multivariate gamma function,
\begin{align*}
\Gamma_m(a) = \int_{H\succ0}e^{-\Tr H}|H|^{a-\frac{m+1}{2}}\,dH = \pi^{m(m-1)/4}\prod_{i=1}^m\Gamma\left(a-\frac{i-1}{2}\right).
\end{align*}
The above holds when $r\le k$. Additionally,
\begin{align*}
\int_{H\succ 0}e^{-\frac{1}{2}\Tr(PH)-\frac{\epsilon}{2}\Tr(QH^{-1})}|H|^{-1}\,dH\sim |P|^{-(m-1)/2}\log\left(\frac{1}{\epsilon}\right).
\end{align*}
By a change of variables, it suffices to show this with $P=I$, and the scaling can be shown by changing variables to
\begin{align*}
H = \begin{pmatrix}
h+u^\top K^{-1} u & u^\top \\ u & K
\end{pmatrix}
\end{align*}
for $h>0$, $u \in \R^{m-1}$ and $K\in\mathbf{S}^{m-1}_{++}$. Thus, setting $k=\floor{n/2}$ and $r=\ceil{n/2}$ we get
\begin{align*}
\int_{A_1\succ 0} e^{-\frac{1}{2} \Tr(A_1(I_r+BB^\top))  - \frac{\epsilon}{2}\Tr(D_{11} A_1^{-1})}|A_1|^{\frac{k}{2}-\frac{r+1}{2}}\,dA_1
\sim
\left(\log\frac{1}{\epsilon}\right)^{\mathbf{1}_{n\text{ odd}}}|I_r + BB^\top|^{-k/2}
\end{align*}
and
\begin{align*}
\int_{A_0\succ0} e^{-\frac{1}{2} \Tr(N_B^\top D N_B A_0^{-1}) -\frac{\epsilon}{2} \Tr A_0}|A_0|^{-(n+1)/2}\,dA_0
\sim
|N_B^\top D N_B|^{-r/2}.
\end{align*}
Finally, we evaluate
\begin{align*}
I_\epsilon(D)\sim c_n\epsilon^{-kr/2}\left(\log\frac{1}{\epsilon}\right)^{\mathbf{1}_{n\text{ odd}}}\int_{\R^{k\times r}} |N_B^\top D N_B|^{-r/2} | I_r+BB^\top|^{-k/2}\,dB.
\end{align*}
Define the column-normalized matrix
\begin{align*}
Q_B = N_B(N_B^\top N_B)^{-1/2} = N_B(I_k + B^\top B)^{-1/2}.
\end{align*}
Then
\begin{align*}
|N_B^\top D N_B| = \left| (N_B^\top N_B)^{1/2}(Q_B^\top D Q_B)(N_B^\top N_B)^{1/2}\right|
= |I_k + B^\top B| |Q_B^\top D Q_B|.
\end{align*}
Since
\begin{align*}
|I_k + B^\top B| = |I_r + BB^\top|,
\end{align*}
the integrand is
\begin{align*}
|N_B^\top D N_B|^{-r/2} | I_r+BB^\top|^{-k/2}
=|Q_B^\top D Q_B|^{-r/2}|I_k + B^\top B|^{-n/2}.
\end{align*}
But $|I_k + B^\top B|^{-n/2}\,dB$ is the invariant measure on $\text{Gr}(k,n)$, which we write as $\nu_{k,n}$. The result is thus proved.

\paragraph{Passing to the distributional limit}
As in the diagonal case, formally, one must check that the convergence $Z_\alpha(W)/C_\alpha\to\Omega(W)$ holds in $L^1(e^{-\ell(W)/T}\,dW)$ to obtain convergence of $p_{\lambda,T}(W)$. For losses satisfying $\ell(W)\ge a\|W\|_F^2-b$, we can find an integrable dominating function. In even dimensions $Z_\alpha(W)/C_\alpha\lesssim\Omega(W)$; in odd dimensions an additional factor $1+|\log(1/s_1(W))|$ suffices (where $s_1$ is the smallest singular value). Both are Gaussian-integrable, so dominated convergence applies.

\subsection{Square ReLU networks integral limit}
\label{app:square-relu}
We examine the integral
\begin{align*}
\int\exp\left(-\frac12 \sum_{e=(u,v)} \bar W_e^2 \alpha^{1-(s_v-s_u)}\right)\,ds
\end{align*}
for $s_v\in \R$. As shown in the main text, the integrand converges to the indicator on the interior of the set $\mathcal{P}=\{s: s_v-s_u\le 1 ~\forall e=(u,v\}$. This set has finite volume, since if $s_v-s_u\le1$ for all nodes in the network, then by induction $-(D-d)\le s_{d,i}\le d$ for all $d\in[D],i\in[n_d]$. To apply the dominated convergence theorem, we show the integrand is bounded by
\begin{align*}
\exp\left(-\frac{1}{2}\left(\min_{e=(u,v)} \bar W_e^2\right) e^{-1+\max_v |s_v| / D}\right)
\end{align*}
(which is non-trivial on the chosen stratum).
Indeed for $\max_v |s_v|>D$, some path from $v$ to the input or output has an edge $e=(u',v')$ such that $s_{v'}-s_{u'}\ge \max_v |s_v|/D$. The bound follows once $\alpha<e^{-1}$, since then $\alpha^{1-(s_{v'}-s_{u'})}\ge \exp(-1+\max_v|s_v|/D)$. For $\max_v |s_v|\le D$, on the other hand, we have the trivial upper bound of 1.

In this case we do not analyze whether $p_{\lambda,T}$, converges, since this point becomes moot under heat death.

\subsection{Square-to-scalar ReLU networks}
\label{app:square-to-scalar-relu}
We consider the ReLU network $f(x)=w_D^\top\sigma(W_{D-1}\cdots\sigma(W_1x))$ with $W_d\in \R^{n\times n}$ for $d<D$ and $w_D \in \R^n$, with $n\ge2$. With the normalization $W_d=P_d\bar W_d P_{d-1}^{-1}$ as described in the main text (with $w_D = P_{D-1}^{-1} \bar w_D$),
\begin{align*}
R(\bar W,\rho) = \frac{1}{2}\left(
\sum_i \rho_{1,i}^2 + \sum_{d=2}^{D-1}\sum_{j,i}(\bar W_d)_{ji}^2\left(\frac{\rho_{d,j}}{\rho_{d-1,i}}\right)^2 + \sum_i \left(\frac{\bar w_{D,i}}{\rho_{D-1,i}}\right)^2
\right)
\end{align*}
and
\begin{align*}
dW = d\omega(\bar W)d\bar w_D\left(\prod_{d=1}^{D-2}\prod_i \rho_{d,i}^{-1}\,d\rho_{d,i}\right)\left(\prod_i \rho_{D-1,i}^{n-2}\,d\rho_{D-1,i}\right).
\end{align*}
Performing the change of variables $\rho_{d,i} = \alpha^{-d/2}s_{d,i}$,
\begin{align*}
\alpha R(\bar W, \rho) = \frac{1}{2}\left(\sum_i s_{1,i}^2 + \sum_{d=2}^{D-1}\sum_{j,i} (\bar W_d)_{ji}^2\frac{s_{d,j}^2}{s_{d-1,i}^2} + \alpha^D \sum_i\frac{\bar w_{D,i}^2}{s_{D-1,i}^2}\right)
\end{align*}
while the measure keeps the same form except for a scaling,
\begin{align*}
dW = \alpha^{-(D-1)n(n-1)/2}d\omega(\bar W)d\bar w_D\left(\prod_{d=1}^{D-2}\prod_i s_{d,i}^{-1}\,ds_{d,i}\right)\left(\prod_i s_{D-1,i}^{n-2}\,ds_{D-1,i}\right).
\end{align*}
Writing $d\mu(s)$ for the $s$-dependent part of the measure above,
\begin{align*}
Z_\alpha(\bar W) = \alpha^{-(D-1)n(n-1)/2} \int e^{-\frac{1}{2}\left(\sum_i s_{1,i}^2 + \sum_{d=2}^{D-1}\sum_{j,i} (\bar W_d)_{ji}^2\frac{s_{d,j}^2}{s_{d-1,i}^2} + \alpha^D \sum_i\frac{\bar w_{D,i}^2}{s_{D-1,i}^2}\right)}\,d\mu(s).
\end{align*}
Only the last term in the integrand contains $\alpha$. Taking it to zero (applying dominated convergence), and provided the integral below is finite, we get $Z_\alpha(\bar W) \sim C_\alpha \Omega(\bar W)$ with
\begin{align*}
C_\alpha = \alpha^{-(D-1)n(n-1)/2},
\qquad
\Omega(\bar W) = \int e^{-\frac{1}{2}\left(\sum_i s_{1,i}^2 + \sum_{d=2}^{D-1}\sum_{j,i} (\bar W_d)_{ji}^2\frac{s_{d,j}^2}{s_{d-1,i}^2}\right)}\,d\mu(s).
\end{align*}
Note the scaling $C_\alpha$ is dominated by the scaling of the heat death stratum from the main text.
To see the integral is finite, note the integrand is upper bounded by keeping only the diagonal terms,
\begin{align*}
\exp\left(-\frac{1}{2}\left(\sum_i s_{1,i}^2 + \sum_{d=2}^{D-1}\sum_{j,i} (\bar W_d)_{ji}^2\frac{s_{d,j}^2}{s_{d-1,i}^2}\right)\right)
\le
\exp\left(-\frac{1}{2}\left(\sum_i s_{1,i}^2 + \sum_{d=2}^{D-1}\sum_{i} (\bar W_d)_{ii}^2\frac{s_{d,i}^2}{s_{d-1,i}^2}\right)\right).
\end{align*}
We obtain an upper bound with decoupled integrals
\begin{align*}
\Omega(\bar W) \le \prod_i \int e^{-\frac{1}{2} \left( s_{1,i}^2 + \sum_{d=2}^{D-1}(\bar W_d)_{ii}^2\frac{s_{d,i}^2}{s_{d-1,i}^2}\right)}\,d\mu_i(s_i)
\end{align*}
where $d\mu_i$ is the $s_i$-dependent part of $d\mu$. Changing variables to $u_{d,i} = s_{d,i}/s_{d-1,i}$, we find the upper bound is equal to
\begin{align*}
    \left(\prod_i\int u_{1,i}^{n-2}e^{-u_{1,i}^2 / 2}\,du_{1,i}\right)\prod_{d=2}^{D-1}\prod_i \int u_{d,i}^{n-2}e^{-(\bar W_d)_{ii}^2 u_{d,i}^2 / 2}\,du_{d,i}.
\end{align*}
This expression is finite, being the product of Gaussian integrals.

\paragraph{Feature sparsity bias}
In terms of nodes $u,v$
\begin{align*}
\Omega(\bar W) = \int \exp\left(-\frac{1}{2} \sum_{\substack{e=(v,u)\\ v\text{ hidden }}} \bar W_{e}^2\left(\frac{s_v}{s_u}\right)^2\right)\,d\mu(s),
\end{align*}
where $s_u=1$ for $u$ a coordinate of the input $x$. Integrating over a hidden node $v$ not in the last hidden layer (assuming at least two hidden layers),
\begin{align*}
\Omega(\bar W) = \int (s_{-v} \text{ term}) \int\exp\left(-\frac{1}{2} \sum_{u\to v} \frac{\bar W_{(u,v)}^2}{s_u^2}s_v^2 -\frac{1}{2} \sum_{v\to u}\bar W_{(u,v)}^2s_u^2\frac{1}{s_v^2} \right)\,\frac{ds_v}{s_v}\,d\mu_{-v}(s_{-v}).
\end{align*}
Suppose that the norm of the outgoing normalized weights $\bar W$ from $v$ is $O(\epsilon)$. To obtain a lower bound on the integral, restrict all other scales to a finite box. On this region,
\begin{align*}
\sum_{u\to v} \frac{\bar W_{(u,v)}^2}{s_u^2} \sim 1
\qquad
\sum_{v\to u}\bar W_{(u,v)}^2s_u^2 \lesssim \epsilon^2.
\end{align*}
By our usual cutoff analysis, the inner integral is lower bounded as $\log(1/\epsilon)$, and thus $\log \Omega(\bar W)\gtrsim\log\log(1/\epsilon)$. This is exactly a sparsity bias for nodes $v$ in non-last hidden layers: the entropy increases as the feature coming from node $v$ becomes unused.

\subsection{Two layer homogeneous network}
\label{app:two-layer-homog}
Consider a two-layer homogeneous-activation network
\begin{align*}
f(x) = \sum_{i=1}^m u_i \sigma\left(v_i^\top x\right)
\end{align*}
for $u_i \in \R$ and $v_i \in \R^n$, and $\sigma:\R\to\R$ with $\sigma(cz) = c^q \sigma(z)$ for $c>0$ (where $q>0$).
This has a scaling symmetry $v_i \mapsto cv_i$, $u_i \mapsto u_i / c^q$. We may thus let
\begin{align*}
r_i = \rho_i^q u_i \qquad \varphi_i = \frac{v_i}{\rho_i} \qquad \rho_i = \|v_i\|,
\end{align*}
so
\begin{align*}
f(x) = \sum_{i=1}^m r_i \sigma\left(\varphi_i^\top x\right).
\end{align*}
Squared-norm regularization on the parameters then takes the form
\begin{align*}
R(u,v) = \frac{1}{2} \sum_{i=1}^m u_i^2 + \|v_i\|^2 
= \frac{1}{2} \sum_{i=1}^m \frac{r_i^2}{\rho_i^{2q}} + \rho_i^2.
\end{align*}
Changing measure gives
\begin{align*}
\prod_{i=1}^m du_i\,dv_i = \prod_{i=1}^m \rho_i^{n-q-1} \,d\rho_i\,dr_i d\omega(\varphi_i)
\end{align*}
where $d\omega$ indicates the uniform measure on the unit $(n-1)$-sphere. If we treat $r$ and $\varphi$ as effective parameters, we get
\begin{align*}
p_{\lambda,T}(r, \varphi) \prod_{i=1}^m dr_i\,d\omega(\varphi_i)
\propto
e^{-\ell(r,\varphi)/T} Z_{\lambda/T}(r,\varphi) \prod_{i=1}^m dr_i\,d\omega(\varphi_i)
\end{align*}
where
\begin{align*}
Z_\alpha(r,\varphi) &= \int_0^\infty e^{-\frac{\alpha}{2}\sum_{i=1}^m \rho_i^{-2q} r_i^2 + \rho_i^2} \prod_{i=1}^m \rho_i^{n-q-1} \,d\rho_i\\
&= \prod_{i=1}^m \int_0^\infty e^{-\frac{\alpha}{2}(\rho_i^{-2q}r_i^2 + \rho_i^2)} \rho_i^{n-q-1}\,d\rho_i.
\end{align*}
Taking the $T\to0$ limit gives
\begin{align*}
R_\text{eff}(r,\varphi) = \frac{1}{2}\inf_\rho\sum_{i=1}^m \frac{r_i^2}{\rho_i^{2q}} + \rho_i^2. 
\end{align*}
Each term in the sum can be minimized independently, with
\begin{align*}
\rho_i = \left(\sqrt{q}|r_i|\right)^{\frac{1}{q+1}}
\quad \Rightarrow \quad
R_\text{eff}(r,\varphi)
= \frac{1+q}{2q^{\frac{q}{q+1}}} \sum_{i=1}^m |r_i|^\frac{2}{q+1} 
\end{align*}
This is minimized at $r=0$.
For example, for $q=1$ (such as for linear or ReLU activations), we get $R_\text{eff}(r,\varphi)=\|r\|_1$.

To obtain the $\lambda\to0$ limit, we analyze the scaling of the integral for a single neuron,
\begin{align*}
I_\alpha(r)=\int_0^\infty e^{-\frac{\alpha}{2}(r^2\rho^{-2q} + \rho^2)} \rho^{n-q-1}\,d\rho,
\end{align*}
where we now write $r,\rho>0$, dropping subscripts for clarity. As $\alpha\to0$, the exponential gives the cutoff
\begin{align*}
\alpha\rho^{-2q} \lesssim 1, \qquad \alpha\rho^2 \lesssim 1,
\end{align*}
that is,
\begin{align*}
\alpha^{\frac{1}{q}} \lesssim \rho^2 \lesssim \frac{1}{\alpha}.
\end{align*}
Thus, changing variables to $y=\alpha\rho^2$, we get $dy=2\alpha\rho \,d\rho$ and
\begin{align*}
I_\alpha(r) &= \int_0^\infty e^{-\frac{1}{2}(\alpha^{q+1}r^2y^{-q}+y)} \left(\frac{y}{\alpha}\right)^{\frac{n-q-1}{2}}\frac{1}{2\sqrt{\alpha y}}\,dy\\
&= \frac{1}{2}\alpha^{-\frac{n-q}{2}} \int_0^\infty e^{-\frac{1}{2}(\alpha^{q+1}r^2y^{-q}+y)} y^\frac{n-q-2}{2}\,dy.
\end{align*}
When $n>q$, we can set $\alpha=0$ in the integral to get
\begin{align*}
I_\alpha(r) \sim  2^\frac{n-q-2}{2} \alpha^{-\frac{n-q}{2}} \Gamma\left(\frac{n-q}{2}\right),
\end{align*}
which is independent of $r$. If $n=q$, then the integral scales like $q^{-1} \log(r^{-2}\alpha^{-q-1}) / 2$, which gives a logarithmic correction vanishing in the renormalized $\alpha\to0$ limit. Finally, if $n<q$, changing variables to $y=r^{2/q}\alpha^{(q+1)/q}z$ gives
\begin{align*}
I_\alpha(r) \sim \frac{1}{q}2^{-\frac{n+q}{2q}}\alpha^{-\frac{q-n}{2q}} \Gamma\left(\frac{q-n}{2q}\right)|r|^{-\frac{q-n}{q}},
\end{align*}
and we obtain the renormalized entropy
\begin{align*}
\log \Omega(r,\varphi) = -\mathbf{1}_{q>n} \frac{q-n}{q} \sum_{i=1}^m \log |r_i|.
\end{align*}
We think of $q$ as usually being small ($1$ for ReLU or $2$ for a quadratic activation), and thus we predict that there is no symmetry-induced implicit bias in neuron magnitude or direction in networks of non-trivial width.

\subsection{Functional entropy}
\label{app:functional-entropy}
\paragraph{Hausdorff measure} Our definition of renormalized entropy over functions requires a measure $df$, ideally one which we can informally think of as uniform. We use a standard construction, by picking a distribution over inputs $x\sim Q$ and considering functions $f$ as members of $L^2(Q)$. On a regular region $S\subseteq \Theta$ on which $\text{rank}(D\pi)=k$, the image $\pi(S)$ is a $k$-dimensional manifold. For a local chart $\phi:z\in\R^k\mapsto f$, the volume form is
\begin{align*}
d\mathcal H^k(f)=\sqrt{{\det}^+ G_Q(z)}\,dz,
\qquad
(G_Q)_{ij}=\langle\partial_i\phi,\partial_j\phi\rangle_{L^2(Q)}.
\end{align*}
Noting $\pi=\phi\circ\iota$ where $\iota:\theta\mapsto z$ take parameters to the effective chart, and applying the coarea formula \citep{simon2018gmt}, give the partition function in the main text.

\paragraph{Two layer homogeneous networks}
Here we build on \cref{app:two-layer-homog} to analyze the entropy $\log\Omega(f)$ with respect to $df$. Recall that $f(x)=\sum_i u_i\sigma(v_i^\top x) = \sum_i r_i \sigma(\varphi_i^\top x)$ for $r_i=u_i\|v_i\|^q$ and $\varphi_i = v_i/\|v_i\|$, where $\sigma$ is the ReLU function. Consider the generic $n>q$ case. We assume there are $n\ge2$ neurons, and the data distribution $Q$ has a finite second moment (and as in the main text, has a density which is positive on a neighborhood of zero).

For the $(r,\varphi)$ parameterization above, the feature derivatives are given by
\begin{align*}
\frac{\partial f(x)}{\partial r_i} = \sigma(\varphi_i^\top x),    \qquad  D_{\varphi_i}f(x)[b_i] = r_i \langle\nabla^S_\varphi \sigma(\varphi^\top x),b_i\rangle
\end{align*}
where $b_i \in T_{\varphi_i}S^{n-1}$ is any infinitesimal change in $\varphi_i$, and
\begin{align*}
\nabla^S_\varphi \sigma(\varphi^\top x) = \sigma'(\varphi^\top x)(I-\varphi\varphi^\top)x
\end{align*}
is the spherical gradient, assuming $\sigma$ is almost everywhere differentiable.
Let $\tilde{K}(\varphi)$ be the Gram matrix of the \emph{unscaled} features,
\begin{align*}
\tilde{K}_{\sigma,Q}(\varphi)= \E_{x\sim Q} \left[ \left(\sigma(\varphi_i^\top x), \sigma'(\varphi_i^\top x) B_i^\top x\right)_{i=1}^m \otimes \left(\sigma(\varphi_j^\top x), \sigma'(\varphi_j^\top x) B_j^\top x\right)_{j=1}^m  \right],
\end{align*}
where $B_i\in\R^{n\times(n-1)}$ are any orthonormal frames for the tangent spaces at $\varphi_i$, meaning $B_i^\top B_i = I_{n-1}$ and $B_i^\top \varphi_i = 0$. (The spectral structure of $\tilde K$ is independent of the choice of $B_i$.) The effective parameter NTK (or rather, its transpose) is given by
\begin{align*}
K(r,\varphi) = D(r) \tilde K_{\sigma,Q}(\varphi) D(r), \qquad D(r)= \text{diag}(1, r_1 I_{n-1},\ldots, 1, r_m I_{n-1}),
\end{align*}
and therefore
\begin{align*}
\sqrt{\det K(r,\varphi)} = \sqrt{\det \tilde K_{\sigma,Q}(\varphi)} \prod_i |r_i|^{(n-1)},
\end{align*}
our assumptions on $Q$ and $n$ guaranteeing positive determinant. This gives the result stated in the main text.

As we stated therein, for ReLU networks, when all $r_i \ne 0$ and $\varphi_i \ne \pm \varphi_j$ for all $i\ne j$, the unordered collection of $(r_i,H_i)$ for hyperplanes $H_i = \{x: \varphi_i^\top x=0\}$ is determined by $f$. This is because $\nabla f$ changes when crossing a hyperplane $H_i$, with magnitude identifying $|r_i|$ and convexity determining the sign. This allows us to more precisely analyze the angular term elided in the main text. First,
\begin{align*}
\Omega(f) = \prod_i |r_i|^{-(n-1)} \sum_{\varphi:\bar\pi(r,\varphi) =f} \frac{1}{\sqrt{\det \tilde K_{\sigma,Q}(\varphi)}}.
\end{align*}
Since there are finitely many choices of orientations of $\varphi_i$ matching the collection of $H_i$, 
\begin{align*}
\log \sum_{\varphi:\bar\pi(r,\varphi) =f}\frac{1}{\sqrt{\det \tilde K_{\sigma,Q}(\varphi)}} = -\frac{1}{2}\log \Delta(f) + O(1)
\end{align*}
where we define the angular term
\begin{align*}
\Delta(f) = \min_{\varphi:\bar\pi(r,\varphi) =f} \det \tilde K_{\sigma,Q}(\varphi).
\end{align*}
Therefore
\begin{align*}
\log \Omega(f) = (n-1) \sum_{i} \log \frac{1}{|r_i|} -\frac{1}{2} \log \Delta(f) + O(1).
\end{align*}
If additionally $Q$ is rotationally invariant, then $\det \tilde K(\varphi)$ depends only on the collection of angles $\arccos(\varphi_i^\top\varphi_j)$. In particular one can show a single vanishing angle of order $\epsilon_{ij}$ contributes $(n+2)/2 ~\log(1/\epsilon_{ij})$ to the entropy.

\subsection{Subleading low-temperature terms and the Menon-Yu entropy}
\label{app:menon-yu-entropy}
\paragraph{Subleading terms}
Consider the square $n$-dimensional linear network $f(x)=UV^\top x$ with parameters $\theta=(U,V)$, and let $W=UV^\top$ with singular value decomposition $W=PSQ^\top$ where $S=\text{diag}(s_1,\ldots,s_n)$. Recall that
\begin{align*}
Z_\alpha(W) \propto \int_{X\succ0} e^{-\frac{\alpha}{2}\Tr(S(X+X^{-1}))}|X|^{-(n+1)/2}\,dX.
\end{align*}
We begin by finding the $T\to0$ (that is, $\alpha\to\infty$) expansion by Laplace's method, including subleading terms. As usual, the leading term is found by noting the minimizer of the exponent, $X=I$. Now we also consider a symmetric perturbation $A=A^\top$ (the tangent space to the positive definite matrices at $I$ being $\text{Sym}(n)$)
\begin{align*}
X = I+\alpha^{-1/2} A \qquad\Longrightarrow \qquad
X^{-1}= I- \alpha^{-1/2}A + \alpha^{-1}A^2 + O(\alpha^{-3/2}).
\end{align*}
The exponent thus has the expansion
\begin{align*}
\frac{\alpha}{2}\Tr(S(X+X^{-1})) = \alpha \Tr(S) + \frac{1}{2}\Tr(SA^2) + O(\alpha^{-1/2}).
\end{align*}
While the integrating measure transforms as
\begin{align*}
|X|^{-(n+1)/2} = 1+ O(\alpha^{-1/2}),\qquad
dX = \alpha^{-n(n+1)/4}\,dA.
\end{align*}
Applying Laplace's method, as $\alpha\to\infty$,
\begin{align*}
Z_\alpha(W) = \alpha^{-n(n+1)/4} e^{-\alpha\|W\|_*} \int_{\text{Sym}(n)} e^{-\frac{1}{2}\Tr(SA^2)}\,dA~(1+o(1)).
\end{align*}
But since $A$ is symmetric
\begin{align*}
\Tr(SA^2) = \sum_i s_i A_{ii}^2 + \sum_{i< j} (s_i+s_j)A_{ij}^2,
\end{align*}
and the Gaussian integral factorizes
\begin{align*}
\int_{\text{Sym}(n)} e^{-\frac{1}{2}\Tr(SA^2)}\,dA=\left(\prod_i\int e^{-s_ih^2/2}\,dh\right)\left(\prod_{i<j}\int e^{-(s_i+s_j)h^2/2}\,dh\right). 
\end{align*}
Up to a constant factor, this is $|W|^{-1/2}\prod_{i<j}\left(s_i+s_j\right)^{-1/2}$, and so
\begin{align*}
Z_\alpha(W) \sim \alpha^{-n(n+1)/4}\exp\left(-\alpha\|W\|_* -\frac{1}{2}\log|W|-\frac{1}{2}\sum_{i<j}\log(s_i+s_j)\right)(1+o(1)).
\end{align*}
The subleading term, describing fluctuations in the fiber, is thus $-\frac{1}{2}\log|W|-\frac{1}{2}\sum_{i<j}\log(s_i+s_j)$.

\paragraph{Relationship to Menon and Yu's entropy}
\citet{menon2025entropy} focus on the \emph{balanced manifold}
\begin{align*}
\mathcal M = \left\{(U,V):U^\top U = V^\top V\right\}.
\end{align*}
The ambient space of $\theta$ is Euclidean with Frobenius inner product, and the balanced manifold inherits the corresponding surface-volume measure $d\mathrm{vol}_{\mathcal M}$. Menon and Yu then consider the piece of the fiber $\pi^{-1}(W)$ containing balanced factorizations
\begin{align*}
    \mathcal O_W=\pi^{-1}(W)\cap\mathcal M.
\end{align*}
Theorem 4 of \citet{menon2025entropy} characterizes the entropy
\begin{align*}
    S_{MY}(W) := \log\mathrm{vol}(\mathcal O_W) = \frac{1}{2} \sum_{i<j} \log(s_i + s_j) + C
\end{align*}
where the volume of $\mathcal O_W$ is also induced by the ambient Euclidean metric. This appears related to, but is not quite, the subleading term we found above. This issue is a change of measure. Our measure over effective parameters is $Z_\alpha(W)\,dW$. The natural measure for Menon and Yu is instead their $d\mathrm{vol}_g(W)$, induced by their metric $g$. This is the metric corresponding to the NTK in our introduction, which turns parameter space gradient flow into a metric gradient flow in $W$. From their equation (3.13),
\begin{align*}
g_W(A,A)=\sum_i \frac{A_{ii}^2}{2s_i} + \sum_{i\ne j} \frac{A_{ij}^2}{s_i+s_j},
\end{align*}
where we write $A$ in the $p_iq_j^\top$ basis: $A_{ij} = p_i^\top A q_j$. As this basis is orthonormal with respect to the Frobenius inner product inducing $dW$,
\begin{align*}
    d\mathrm{vol}_g(W) = \left(\prod_i \frac{1}{2s_i}\right)^{1/2}\left(\prod_{i\ne j} \frac{1}{s_i+s_j}\right)^{1/2}\,dW \propto|W|^{-1/2}\prod_{i<j}(s_i+s_j)^{-1}\,dW.
\end{align*}
Combining terms and defining the appropriate constant $C_\alpha$ we find that
\begin{align*}
Z_\alpha(W)\,dW \sim C_\alpha \exp\left(-\alpha\|W\|_* + S_{MY}(W)\right)\, d\mathrm{vol}_g(W).
\end{align*}

\paragraph{Interpretation as balanced parameter volume}
Following our main text, we define the measure in effective parameter space counting balanced representatives,
\begin{align*}
\nu_\mathcal M(B) = \int_{\pi^{-1}(B)\cap\mathcal M} d\mathrm{vol}_\mathcal M(\theta).
\end{align*}
(No regularization is needed here, since $\mathcal O_W$ is compact.) Menon and Yu's Theorem 8 states that the restriction $\pi|_\mathcal M$ is a Riemannian submersion when the image is given their metric $g$.  By the coarea formula, it follows that
\begin{align*}
\nu_\mathcal M(dW) =  \mathrm{vol}(\mathcal O_W) d\mathrm{vol}_g(W)= e^{S_{MY}(W)} d\mathrm{vol}_g(W).
\end{align*}
What our argument above then shows is that
$$Z_\alpha(W) \, dW \sim C_\alpha \exp(-\alpha\|W\|_*)\,\nu_\mathcal M(dW).$$
That is, at low temperature, fluctuations in $W$ are governed exactly by movement in the balanced manifold.

\section{Experiments}

\subsection{Diagonal networks}
\label{app:diag-experiment}
Our diagonal experiments use depth two, with $u,v\in\mathbb R^2$, effective coefficients $w=u\odot v$, target $w_*=(1,-1)$, and loss $\ell(w)=\|w-w_*\|_2^2/2$. We use temperature $T=0.2$ and independent Gaussian initialization of each parameter with mean zero and standard deviation $s_0=0.5$.

For each $\lambda\in\{0,0.003,0.01,0.03,0.1\}$, we evolve 512 independent trajectories to $t=10^4$. Integration uses a symmetric splitting scheme with nominal step $h=0.1$: with the other layer fixed, we sample the exact conditional Ornstein–Uhlenbeck transition, applying successive updates of durations $h/2$, $h$, and $h/2$ to $u$, $v$, and $u$, respectively. Steps are shortened to reach observation times exactly. Although the conditional transitions are exact, the finite-step transient dynamics approximate the continuous-time Langevin process.

We record one coefficient vector per trajectory at $t=0$ and $t_k=10^{-1+k/4}$ for $k=0,\ldots,20$. The main histogram and CDF panels show $t=10^{-1/4}\approx0.56234$, $10^3$, and $10^4$. These are transient ensembles initialized at $t=0$; no burn-in or temporal thinning is applied.

For positive regularization, the predicted coordinate densities are
\[
p_{\lambda,T}(w_j)\propto
\exp\!\left[-\frac{(w_j-w_{*,j})^2}{2T}\right]
K_0\!\left(\frac{\lambda}{T}|w_j|\right),
\]
normalized numerically by quadrature. For the effective regularization comparisons, we evaluate this expression at the prescribed heuristic effective precision
$\alpha_{\mathrm{eff}}(t)=(s_0^2+2Tt)^{-1}$, equivalently $\lambda_{\mathrm{eff}}(t)=T\alpha_{\mathrm{eff}}(t)$. We also compare with the weak-regularization equilibrium limit, whose coordinate marginals are $\mathcal N(w_{*,j},T)$.

\subsection{Linear networks}
\label{app:linear-experiment}
We use a two-layer square linear network \(W=UV^\top\), with \(U,V\in\mathbb R^{4\times4}\), target \(W_*=I_4\), loss \(\ell(W)=\|W-W_*\|_F^2/2\), and temperature \(T=0.2\). Initial parameter entries are independent \(\mathcal N(0,s_0^2)\) variables with \(s_0=0.5\). Regularized experiments add \(\lambda(\|U\|_F^2+\|V\|_F^2)/2\) to the potential. We simulate 2,048 independent trajectories for \(\lambda=0\) and 512 for each \(\lambda\in\{0.003,0.01,0.03,0.1\}\), through \(t=10^4\). Snapshots are recorded at logarithmically spaced times, with four intervals per decade starting at \(t=0.1\). The main distribution panels show \(t\approx0.56234,10^3,10^4\), using one output per trajectory at each time.

We integrate the parameter-space Langevin dynamics using symmetric splitting with step \(h=0.1\). With one matrix fixed, the potential is quadratic in the other, allowing its conditional Ornstein–Uhlenbeck transition to be sampled exactly. Each full step comprises a half-step update of one matrix, a full-step update of the other, and a second half-step update of the first. These conditional updates preserve the positive-\(\lambda\) Gibbs measure in exact arithmetic; finite-step errors remain in the transient dynamics.

The unregularized snapshots are compared with the weak-regularization prediction
\[
\lim_{\lambda\downarrow0}p_{\lambda,T}(W)
\propto e^{-\ell(W)/T}\Omega(W),
\]
the normalized loss-only Gaussian law \(W=W_*+\sqrt T\,\Xi\), where \(\Xi\) has independent standard Gaussian entries, and the heuristic reference \(p_{\lambda_{\mathrm{eff}}(t),T}\), with
\[
\alpha_{\mathrm{eff}}(t)=\frac{1}{s_0^2+2Tt},
\qquad
\lambda_{\mathrm{eff}}(t)=T\alpha_{\mathrm{eff}}(t).
\]
We generate independent samples from the weak-regularization law using matrix-beta augmentation and rejection sampling. Finite-effective-regularization references use Gaussian conditioning and rejection sampling, with a Schur-coordinate proposal at small regularization. The weak-limit and loss-only references each contain 65,536 samples. Effective-regularization references contain 65,536 samples at the displayed distribution times and 4,096 at the remaining times used in the Kolmogorov-Smirnov analysis. The separate fixed-positive-\(\lambda\) comparisons use one endpoint from each of 4,096 independent conditional-Gibbs chains after 10,000 sweeps; these are finite-burn-in equilibrium estimates.

Reference-density curves use Gaussian smoothing with a Silverman-rule bandwidth and reflection at zero. Empirical CDFs and KS statistics use unsmoothed samples. The pointwise 95\% KS intervals use Dvoretzky–Kiefer–Wolfowitz bounds accounting for both trajectory and reference sampling uncertainty.

\subsection{ReLU networks}

\label{app:heat-death-experiment}

We simulate a fully connected, bias-free ReLU network with scalar input, three hidden layers of width three, and a linear scalar readout, giving 24 trainable weights. All weights are initialized independently from \(\operatorname{Uniform}(-\sqrt{3}/2,\sqrt{3}/2)\), with standard deviation \(0.5\). The training inputs are \(x=-1,+1\), with respective targets \(0.5,1\), and the loss is
\[
L(\theta)=\ell(f_\theta)
=\frac14\left[\bigl(f_\theta(-1)-0.5\bigr)^2+
\bigl(f_\theta(1)-1\bigr)^2\right].
\]
For each \(\lambda\in\{0,0.003,0.01,0.03,0.1\}\), we evolve 512 independent Langevin trajectories under the potential \(L(\theta)+\lambda\|\theta\|_2^2/2\), at constant temperature \(T=0.2\), from initialization through \(t=320\). We use adaptive Euler–Maruyama integration with maximum step size \(10^{-3}\), curvature-safety parameter \(0.0125\), and drift-increment parameter \(0.0025\). Observations are recorded at \(t=0,2,\ldots,320\).

A neuron is dead when its activation vanishes identically over the input domain, and the network is dead when \(f_\theta\equiv0\). For this scalar-input, bias-free positively homogeneous architecture, activations and outputs at \(-1,+1\) determine their values on the two input rays; checking these two inputs therefore characterizes the global death events. The heatmaps show each hidden neuron's empirical death probability for the unregularized ensemble at \(t=0,20,320\). The time-dependent probability curves show
\[
\widehat p_0(t,\lambda)
=\frac1{512}\sum_{r=1}^{512}
\mathbf1\!\left\{f_{\theta_t^{(r)}}\equiv0\right\}
\]
for all five regularizations. The accompanying mean-loss curve shows the unregularized ensemble average of \(\ell(f_{\theta_t})\). Reported losses exclude the regularization term. Initially dead networks are retained, and death is not treated as absorbing. Shaded regions are pointwise 95\% Wilson intervals for network death probability and nominal pointwise 95\% Student-\(t\) intervals for ensemble mean loss.

The probability-versus-\(\lambda\) panel shows the equilibrium prediction \(\Pr_{\lambda,T}(f=0)\), separately from the finite-time trajectory estimates. For \(\lambda>0\), we express the Gibbs law as the Gaussian weight distribution \(\mathcal N(0,(T/\lambda)I_{24})\) tilted by \(e^{-\ell(f_\theta)/T}\). We integrate the three Gaussian readout weights analytically and evaluate the remaining 21-dimensional hidden-weight expectation using eight independently scrambled Sobol rules with \(2^{21}\) nodes each. The curve uses 601 regularization values on \([0,0.1]\), with common quadrature nodes across positive regularizations and no fitting to the Langevin data. Its value at \(\lambda=0\) denotes the theoretical weak-regularization limit \(\lim_{\lambda\downarrow0}\Pr_{\lambda,T}(f=0)=1\), rather than a normalizable unregularized parameter-space Gibbs distribution.

\end{document}